\documentclass[lettersize,journal]{IEEEtran}
\usepackage{amsmath,amsfonts}
\usepackage{algorithmic}
\usepackage{algorithm}
\usepackage{array}
\usepackage{textcomp}
\usepackage{stfloats}
\usepackage{url}
\usepackage{verbatim}
\usepackage{graphicx}
\usepackage{cite}
\usepackage{subfigure}
\usepackage{multirow}
\usepackage{color}
\usepackage{soul}
\usepackage[dvipsnames]{xcolor}
\usepackage{amsmath}
\begin{document}

\title{Vision Language Modeling of Content, Distortion and Appearance for Image Quality Assessment}

\author{Fei Zhou, Tianhao Gu, Zhicong Huang, and Guoping Qiu
\thanks{
This work was supported in part by the National Natural Science Foundation of China under Grant 62271323 and U22B2035,  in part by Guangdong Basic and Applied Basic Research Foundation under Grant  2023A1515012956 and 2023B1212060076, and in part by the Shenzhen Research and Development Program under Grant JCYJ20220531102408020 and KJZD20230923114209019.
(\textit{Corresponding author: Guoping Qiu.})}
\thanks{Fei Zhou is with College of Electronic and Information Engineering, Shenzhen University, Shenzhen 518060, China and also with Guangdong Provincial Key Laboratory of Intelligent Information Processing, Shenzhen 518060, China. (e-mail: flying.zhou@163.com).

Tianhao Gu is with College of Electronic and Information Engineering, Shenzhen University, Shenzhen 518060, China, and also with the SZU-AFS Joint Innovation Center for Artificial Intelligence Technology, Shenzhen 518000, China. (e-mail: iceurocha@gmail.com).

Zhicong Huang is with College of Electronic and Information Engineering, Shenzhen University, Shenzhen 518060, China, also with the Guangdong-Hong Kong Joint Laboratory for Big Data Imaging and Communication, Shenzhen 518060, China. (e-mail: hzc10197@qq.com).

Guoping Qiu is with 
School of Computer Science, The University of Nottingham Ningbo China,  Zhejiang 315100, China, and also with the School of Computer Science, University of Nottingham, Nottingham NG8 1BB, UK. (e-mail: guoping.qiu@nottingham.ac.uk).
}
}

\maketitle
\begin{abstract}
The visual quality of an image is confounded by a number of intertwined factors including its semantic content, distortion characteristics and appearance properties such as brightness, contrast, sharpness, and colourfulness. Distilling high level knowledge about all these quality bearing attributes is crucial for developing objective Image Quality Assessment (IQA). 
While existing solutions have modeled some of these aspects, a comprehensive solution that involves all these important quality related attributes has not yet been developed. In this paper, we present a new blind IQA (BIQA) model termed Self-supervision and Vision-Language supervision Image QUality Evaluator (SLIQUE) that features a joint vision-language and visual contrastive representation learning framework for acquiring high level knowledge about the images semantic contents, distortion characteristics and appearance properties for IQA. For training SLIQUE, we have developed a systematic approach to constructing a first of its kind large image database annotated with all three categories of quality relevant texts. The Text Annotated Distortion, Appearance and Content (TADAC\footnote{TADAC database will be made publicly available.}) database has over 1.6 million images annotated with textual descriptions of their semantic contents, distortion characteristics and appearance properties. The method for constructing TADAC and the database itself will be particularly useful for exploiting vision-language modeling for advanced IQA applications. Extensive experimental results show that SLIQUE has superior performances over state of the art, demonstrating the soundness of its design principle and the effectiveness of its implementation.  
The link to the TADAC dataset is https://ieee-dataport.org/documents/tadactext-annotated-distortion-appearance-and-content
\end{abstract}
\vspace{1em} 
\noindent\footnotesize
© 2024 IEEE. Personal use of this material is permitted. Permission from IEEE must be obtained for all other uses, in any current or future media, including reprinting/republishing this material for advertising or promotional purposes, creating new collective works, for resale or redistribution to servers or lists, or reuse of any copyrighted component of this work in other works.
\section{Introduction}
Developing Image Quality Assessment (IQA) models that correlate well with human perceptual judgments is difficult. No-Reference Image Quality Assessment (NR-IQA), also known as blind IQA (BIQA), is more useful in real-world scenarios since it does not require any reference to quantify the  human-perceivable image quality. However, lacking reference information also makes developing BIQA models more challenging. 

Although deep learning-based BIQA methods have significantly advanced state of the art, the key challenge of lacking large labeled training dataset still remains. Existing databases are generally inadequate in size.
A common strategy to overcoming this difficulty is to train a model using synthetic images with artificial distortion and then regress the model to a small-scale target database. However, image distortions in the real world are complex and it is very difficult to simulate all possible real world scenarios. Another approach is to use self-supervised learning (SSL) \cite{chen2020simple,he2020momentum,ijcai2023p188} to overcome the lack of sufficient training data, as SSL is able to leverage large amount of unlabeled data. 
The proposed method \cite{synmos} synthesizes perceptually aligned quality scores by fusing multiple Full-Reference IQA (FR-IQA) metrics, calibrating with human evaluations, and generating Synthetic Quality Benchmark (SQB) scores, enabling robust training of models for BIQA without extensive human labeling. A much more fundamental challenge in BIQA is that human perception of quality can vary with the content of the image \cite{su2020blindly,topiq}, thus image quality is not only affected by distortion but is also closely tied to image content. Therefore content and distortion \textbf{confound} the image quality assessment problem. The challenge in developing BIQA models lies in the difficulty of distinguishing genuine high-frequency image features from distortions caused by compression artifacts, as these can appear similar.

\begin{figure*}[htbp]
	\centering
	\subfigure[Blur]{
\includegraphics[width=0.23\textwidth]{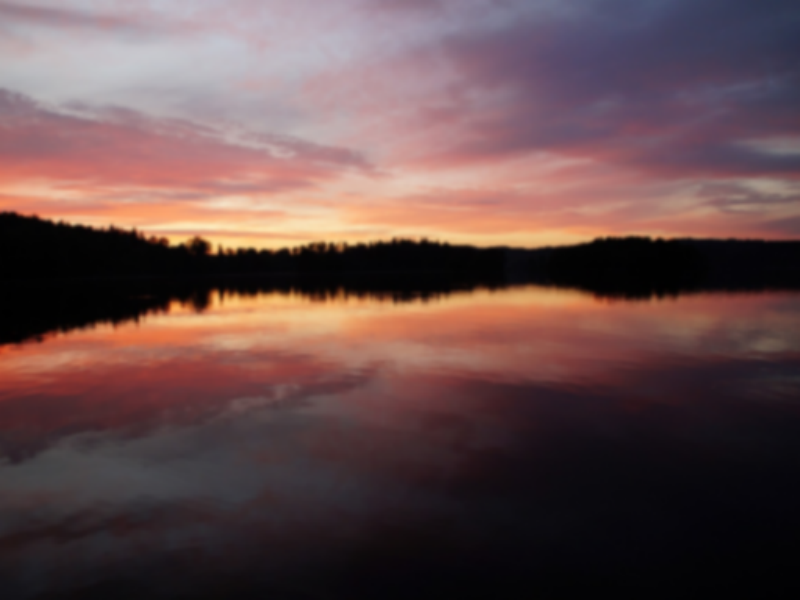}}
\subfigure[Noise]{
\includegraphics[width=0.23\textwidth]{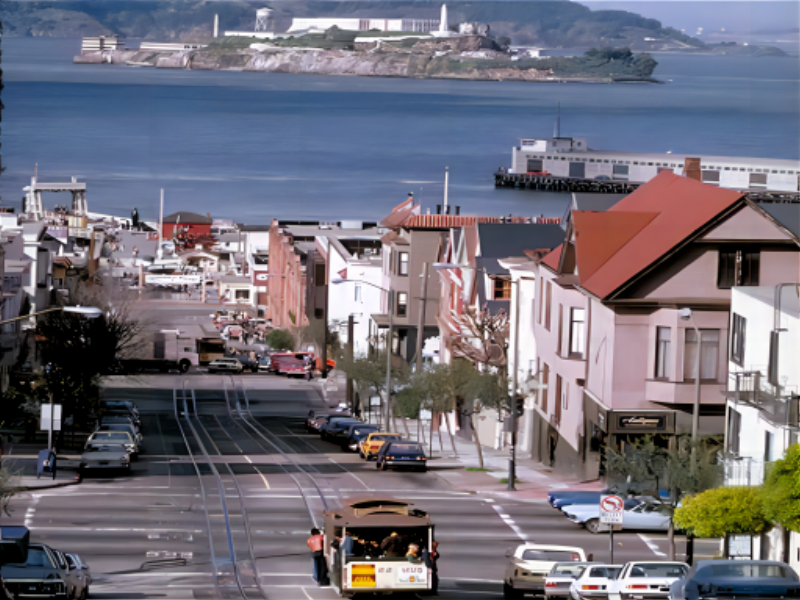}}
\subfigure[Underexposure]{
\includegraphics[width=0.23\textwidth]{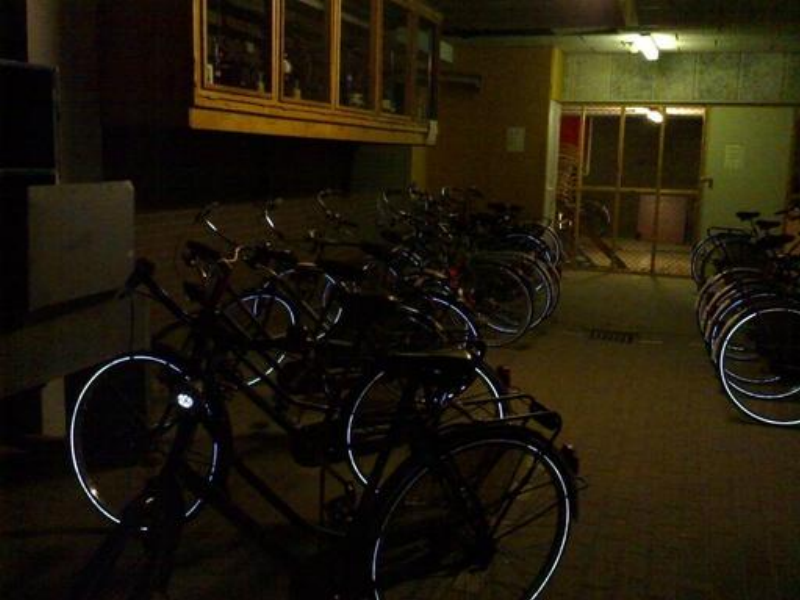}}
\subfigure[Overexposure]{
\includegraphics[width=0.23\textwidth]{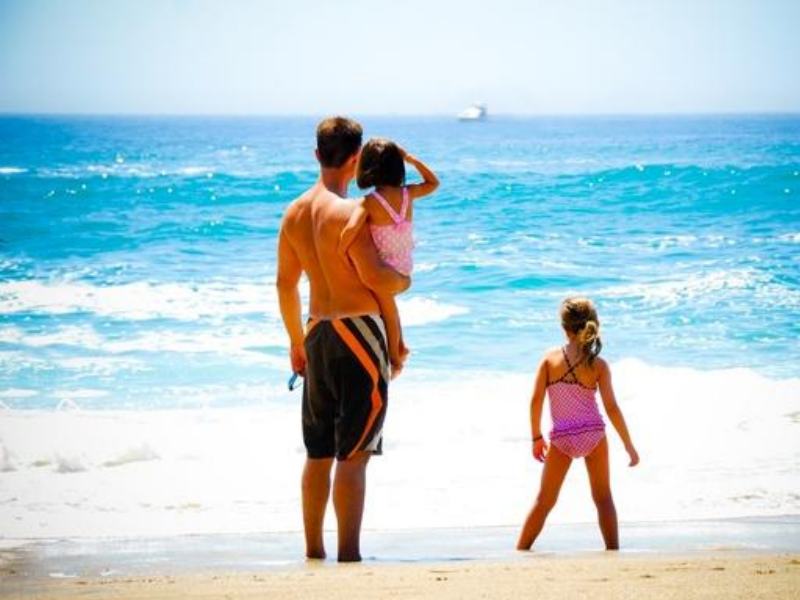}}

\caption{
Example images containing synthetic (a) (b) and real (c) (d) distortions. Optimal when zoomed in.}
\label{Introduction_img}
\end{figure*}
Human have the ability to effortlessly recognize the scene and objects depicted in an image, almost unaffected by distortions. Although it is still unknown how the brain processes content and distortion, it is reasonable and useful to assume, conceptually at least, that it processes the content and the distortion with separately mechanisms.
Fig. \ref{Introduction_img} shows some example images of different contents and distortions. Taking Fig. 1(b) for example, we first notice the scene (city) and then on close inspection we see the image is noisy. Other images can also be separated into content and distortion.
Based on these examples, it seems to make reasonable sense to regard visual quality assessment as consisting of recognising the semantic content of the image and distinguishing the noise or artifacts. A third aspect that will affect the perception of image quality is the image's appearance characteristics such as brightness, contrast, sharpness, and colorfulness. A bright image and a dark image will be perceived as having different quality even if the semantic content and the distortion are the same. Similarly the same image displayed in colour and in gray-scale will have different perceived qualities.
Therefore, a combination of at least three factors, semantic content, distortion and appearance, will affect the perveived visual quality of the image.

Therefore, the challenge of developing BIQA models is how to encode the \textbf{semantic content}, \textbf{distortion} and \textbf{appearance} appropriately and relate them to the final perceived image quality. 
In this paper, we have developed a new model termed Self-supervision and Vision-Language supervision Image QUality Evaluator (SLIQUE) which models the semantic content, distortion and appearance together to assess an image's visual quality. To enable the development of SLIQUE, we contribute a large Text Annotated Distortion, Appearance and Content (TADAC) image database. The database is the first of its kind specifically designed to exploit vision-language modeling for IQA research. The key contributions of this paper are as follows: 
\begin{itemize}
\item We present the Self-supervision and Vision-Language supervision Image QUality Evaluator (SLIQUE) which models the image semantic contents, distortions and appearances together to assess the visual quality of images. SLIQUE exploits vision-language modeling and self-supervised deep learning to distill high level knowledge about the image semantic contents, distortion characteristics and appearance properties together in the BIQA model to provide accurate and reliable assessment of the visual quality of images.

\item We have developed a systematic method for constructing large text annotated image databases designed for exploiting vision-language modeling for image quality assessment and present the Text Annotated Distortion, Appearance and Content (TADAC) database containing over 1.6 million images annotated with texts about their semantic contents, distortion characteristics and appearance properties. We used existing labels or automatic image captioning to annotate the semantic content, designed a list of suitable textual phrases for describing the distortion characteristics, and developed automatic algorithms for computing the appearance properties and annotated these properties with suitable textual descriptions. The database is the first of its kind that is annotated with all three types of quality relevant texts to enable the learning of high level knowledge about all possible factors affecting image quality. TADAC has enabled the development of the first BIQA model (SLIQUE) that jointly models semantic content, distortion and appearance. We will make TADAC publicly available.
\end{itemize}

\begin{figure*}[t]
\centering

\includegraphics[width=1\textwidth]{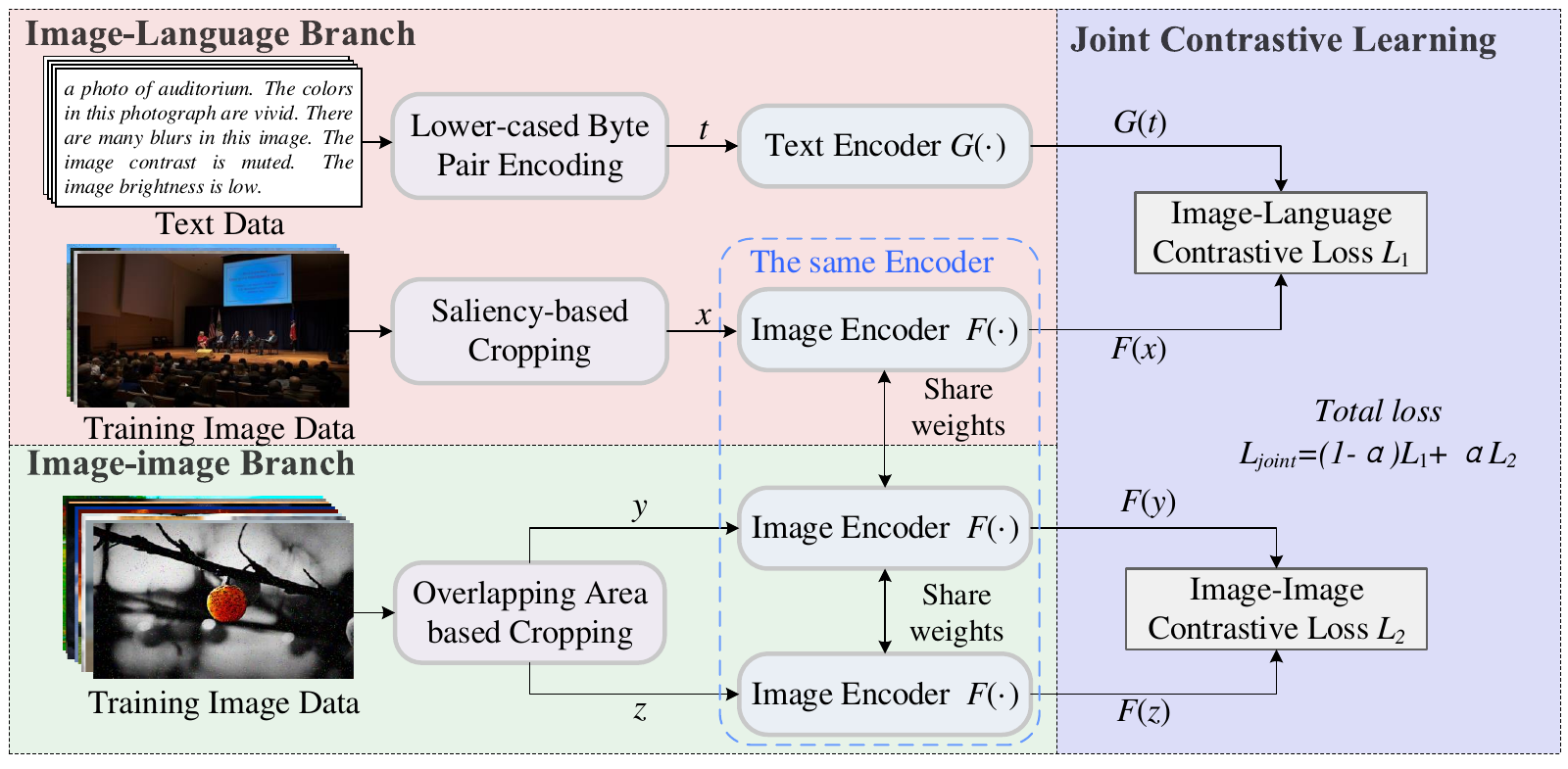}

\caption{Training the image encoder of SLIQUE. The Image-Language Branch performs vision-language constrastive learning for aligning text labels and image contents and the Image-Image Branch carries out self-supervised visual learning. The objective is to train the image encoder for extracting discriminative image features that capture all categories of quality relevant image attributes. Note all three image encoders in the diagram are identical triplets. The purpose of the saliency-based cropping module is to use visual saliency for cropping a high resolution sub-image containing the maximum amount of visual information during training (rather than using the whole image which can be too large). After the system is trained, only the image econder is used for extracting image features for predicting the image quality, see Section \ref{regression}.}

    \label{SLIQUE_Model}

\end{figure*}

\section{Related Work} 
\label{relatedwork}
\subsection{Blind Image Quality Assessment }
Traditional methods mainly focus on designing hand-crafted features as the representation to predict the quality scores. Among them, models based on natural scene statistics (NSS) are widely studied,such as NIQE\cite{mittal2012making}. Another kind of techniques is based on visual codebooks. 

Recently, the success of many deep learning-based computer vision tasks has inspired a number of BIQA models based on deep networks. However, lacking large labeled training dataset is one of the major challenges.
One of the strategies is to use various approaches to augment the training datasets. 
PaQ-2-PiQ \cite{ying2020patches} introduces a large dataset that includes images and patches to train IQA models. MUSIQ \cite{ke2021musiq} uses multi-scale data for training.
The model in \cite{bosse2017deep} predicts the overall quality score of the image by combining scores of image patches, and the model in \cite{zhu2020metaiqa} adopts meta-learning to achieve the training.
The DB-CNN model \cite{zhang2018blind} uses synthetic and authentic distortion images to finetune a dual-stream pretrained network. The staircase structure model in \cite{sun2023blind} integrates feature from intermediate layers into final feature representation through mixed database training strategy. The model in \cite{wang2023blind} implements graph attention network to aggregate features decomposed by fully-connected layers. The model in \cite{zhang2023blind} have developed a multi-task learning method where they apply language-supervised multi-task learning to assess image quality, recognize scenes, and identify distortions. By using textual descriptions of distorted images, they create image-text pairs to finetune the CLIP \cite{radford2019language} model. GRepQ \cite{srinath2024learning} learning  low-level quality features agnostic to distortion types, Further, fine-tuning one such model to extract high-level image
quality information through simple relevant text prompts. 

\subsection{Vision-Language Contrastive Learning }
Vision-language contrastive learning is a mechanism that combines vision and language understanding. It learns to associate images and their textual descriptions in a shared embedding space. Through a contrastive loss function, the embedding textual vector and the visual vector of the same image are encouraged to be closer, while those of different images are pushed farther apart. 
Vision-language contrastive learning bridges the gap between vision and language, opening up new possibilities in image-text analysis. It is commonly used in multimodal tasks, including text-driven image manipulation \cite{gabbay2021image,xu2022predict}, image captioning \cite{hessel2021clipscore}, object detection \cite{zhong2022regionclip,shi2022proposalclip}, and semantic segmentation \cite{rao2022denseclip,zhou2022extract}. The methods in\cite{yang2022vision,uniclip} introduce a triple contrastive learning framework for vision-language pre-training, leveraging both image-text and image-image to enhance cross-model representation learning. The majority of aforementioned works focus on understanding the high-level semantic of images through vision-language contrastive learning. The application of vision-language contrastive learning to learn image representations for BIQA has just started to attract the attention of researchers \cite{wang2023exploring} where 
simple abstract textual descriptions like “good photo” and “bad photo” are used to fine-tune the pre-trained CLIP model \cite{radford2021learning}. However, as visual quality involves multiple influencing factors, it is insufficient to use simple abstract textual descriptions in providing enough textual information for image quality perception. 
Our method addresses the above limitations by designing and adopting text that is more diverse and related to image quality. We embeds text information into the model using a language supervision contrastive learning approach, enabling SLIQUE to more accurately extract image quality features.

\subsection{Visual Self-supervised Learning}
Visual self-supervised learning is also a potential method for overcoming the problem of lacking labeled data.
There are two main approaches: generative-based \cite{He_2022_CVPR} and contrastive-based \cite{he2020momentum,ijcai2023p188}. In BIQA, self-supervised learning has already been utilized.
SAWAR \cite{10196496} applies a self-supervised generative-based learning architecture of collaborative autoencoding, using content autoencoder and distortion autoencoder to represent content and distortion respectively.
CONTRIQUE \cite{madhusudana2022image} employs visual contrastive learning by setting positive and negative samples through the classification of distortion type or distortion severity. Re-IQA \cite{saha2023re} adopts a dual-stream pre-trained network, with one stream using an ImageNet \cite{deng2009imagenet} to train a content encoder and the other stream using distortion dataset to train a distortion encoder.
SPIQ \cite{chen2022spiq} exploits a sample pairing strategy, where sub-images that are distant from each other within an image are defined as negative samples, while those that are closer are defined as positive samples. QPT-Resnet50 \cite{Zhao_2023_CVPR} introduces a distortion degradation space to generate a variety of complex distortion images which are used in visual contrastive learning. In this paper, we not only employ visual contrastive learning but also apply vision-language learning to provide language supervision knowledge together to predict image quality.

\section{SLIQUE} 
\label{SLIQUE}

Our new model is termed Self-supervision and Vision-Language supervision Image QUality Evaluator (SLIQUE) and the training of its major part, the image encoder is shown in Fig. \ref{SLIQUE_Model}. The design of the SLIQUE architecture and training algorithm addresses the major challenges in BIQA as discussed previously. First, it employs self-supervised learning to exploit large amount of readily available unlabeled images. Second, it uses language-vision contrastive learning to embed high level quality relevant knowledge into the model. Third, it constructs a large Text Annotated Distortion, Appearance and Content (TADAC) image database for training the SLIQUE such that it can acquire quality relevant high level knowledge to accurately predict the visual quality of images.

\subsection{Joint Contrastive Learning}
The SLIQUE architecture in Fig. \ref{SLIQUE_Model} has an image-language branch and an image-image branch, and each of the two branches has a two-stream architecture to achieve the joint contrastive learning. 

\textit{Image-language branch}: 
This branch consists of a language encoder $G$ and an image encoder $F$. The encoder $G$ transforms the input textual description $t$ to an embedding feature $G(t)$. Similarly, $F$ transforms the input image $x$ to a feature $F(x)$. Note that $G(t)$ and $F(x)$ have the same dimension by design so that the similarity between them can be readily calculated. The training objective of the image-language branch is to match images with their textual labels which contain descriptions of image content, distortion and appearance information. Specifically, the InfoNCE loss \cite{oord2018representation} is employed to constrain the image-language branch: 

\begin{equation}
\mathcal{\textit{l}}=-\log \frac{\exp \left(q \cdot k_{+} / \tau\right)}{\sum_{i=0}^N \exp \left(q \cdot k_i / \tau\right)},
\end{equation}
where $q$ is an encoded query, $k_i$ is the $i$-th encoded key of a dictionary consisting of \textit{N} samples, and $k_+$ is a single key that matches $q$. And $\tau$ is a temperature hyper-parameter.

\begin{multline}
\mathrm{\textit{L}}_1 = \sum_i l\biggl(\ldots, F(x_n^i), G(t_n^i), \ldots \biggm| \left<x_n^i, t_n^i\right> \in \mathcal{B}_i, \\
1 \leq n \leq N \biggr),
\end{multline}
where $l(\cdot)$ is defined in Equation (1), $\mathcal{B}_i$ denotes the $i$-th training batch consisting of 1 positive sample pair and $N$-1 negative sample pairs, and $\langle x_n^i, t_n^i\rangle$ is the $n$-th $(1 \leq n \leq N)$ sample pair of image-language in $\mathcal{B}_i$. 
 
The pairing strategy can be found in Section \ref{Pairing Strategy}. Based on Equation (2), the image-language branch should be able to capture the intrinsic correlation between visual and textual representations.

 Before sending the input text to the language encoder, we represent it by the lower-cased byte pair encoding (BPE). Another common practice is resizing the input image to a fixed size. However, the resizing operation generally alters the perceptual quality of images. Thus, during training the encoders in Fig. 2, we crop a sub-image from the input image instead of resizing. Cropping is based on the visual saliency of the input. Specifically, a saliency map is first predicted from the input image by hierarchical knowledge embedding \cite{9171494}. Subsequently, a sub-image with a fixed size is cropped at the position where the sum of saliency values within the sub-image is the maximum. As will be shown in ``Implementation Details" part of Section IV, during the training phase of joint contrastive learning, the cropped sub-image covers a considerable part of the input image. Thus, it is reasonable to assume that the image information are well preserved in the cropped images.  
 
\textit{Image-image branch}: 
This branch is similar to the above image-language branch. The main difference is that no language encoder is required here. Instead, two input images $y$ and $z$ are transformed concurrently by the image encoder $F$ into a feature space. The two image encoders in this branch are a pair of identical twins which is exactly the same as the image encoder in the image-language branch. In other words, the three ``Image Encoder" modules illustrated in Fig. 2 are identical triplets and share the same parameters all the time. This design is important, since it ensures the output features from the two branches are in the same embedding space.

Another difference is that training samples prepared for this branch are image-image pairs rather than image-language ones. The training objective of image-image branch is to enhances the model's ability to capture subtle visual details that textual descriptions may miss. The constraint on the image-image branch can be expressed as Equation (3).
\begin{multline}
\mathrm{\textit{L}}_2 = \sum_j  l\biggl(\ldots, F(y_m^j),F(z_m^j),\ldots \biggm|\left<y_m^j, z_{m}^j\right> \in \mathcal{A}_j, \\
1 \leq m \leq M \biggr),
\end{multline}
where $\mathcal{A}_j$ denotes the $j$-th training batch consisting of 1 positive sample pair and $M$-1 negative sample pairs and $\langle y_m^j, z_m^j \rangle$ is the $m$-th $(1 \leq m \leq M )$ sample pair of image-image in $\mathcal{A}_j$. Here we adopt the overlapping area (OLA) based cropping \cite{saha2023re} to ensure that two sub-images in a positive sample pair are with similar quality. Details of the pairing strategy can be found in Section \ref{Pairing Strategy}. 

\textit{Joint training}: 
To determine the learnable parameters in the encoders $G$ and $F$, we do not directly employ or fine-tune publicly released models\cite{mu2022slip} that have already been trained for other tasks. 
Instead, we jointly optimize the two branches from scratch based on contrastive learning:              

\begin{equation}
L_{joint}= ({1-\alpha}) L_{1} + \alpha L_{2},
\end{equation}
where $\alpha$ is the parameter to balance the image-image branch and the image-language branch. 
Notably, $L_1$ and $L_2$ have the same value range, as both of them are based on the InfoNCE loss. Thus, the role of the parameter $\alpha$ is to weight two terms rather than scaling either of them. After minimizing Equation (4), the encoder parameters are frozen. The image encoder $F$ is used to extract image features which are then fed to a simple regression neural network to predict the image quality. 
\subsection{Regression to Quality Score}
\label{regression}

Following the suggestion in \cite{madhusudana2022image}, we employ the ridge regression, a regularized linear regressor, to map the encoded image feature \textbf{h} to a visual quality score $y$:
\begin{equation}
{y = \mathbf{w} \cdot \mathbf{h},}
\end{equation}
where $w$ is the regression vector estimated by
\begin{equation}
{\mathbf{w} = \underset{\mathbf{w}}{\text{argmin}} \sum_k (GT_k - y_k)^2 + \lambda\\|\mathbf{w}\|^2,}
\end{equation}
{where $GT_k$ and $y_k$ are the MOS and the predicted score of the $k$-th image in the training set respectiveily, and $\lambda$ is a regularization constant ($\lambda$ is empirically set to 0.2 in this work).}
Note that given an image to be assessed, its textual description may be not comprehensive or even not available at all in real-world scenarios. Thus, the language encoder is excluded in the regression, and we do not require the textual description of an image to achieve the quality assessment in the testing.

\subsection{Construction of database}
In order to train the joint contrastive representation learning module, we have constructed a large Text Annotated Distortion, Appearance and Content (TADAC) image database. Each image in TADAC is annotated with three types of textual descriptions indicating the semantic contents, the distortion characteristics and the appearance properties respectively. 

\textit{Image Data Preparation}: Images with diverse contents and various distortions are required to train the joint contrastive learning framework. The data should not be biased towards a particular type of image quality, thereby avoiding potential shortcomings in generalization performance. Thus, we collect images with both synthetic distortions and authentic distortions from multiple image databases. Notably, no MOS annotation is required for the collected images. 

For synthetic distortions, we include 700,000 distorted images from the KADIS dataset \cite{deepfl-iqa}, which contains 25 distortion types and each type has 5 levels of severity. Besides, the 140,000 pristine images from this dataset are also included, resulting in a total of 840,000 images. Although these images have no MOS annotations, they have already been annotated by semantic labels (scene and object) and distortion labels (type and level). 


In addition to synthetic distortion images, we collected 809,850 images with authentic (real) distortion from five datasets: 255,000 from AVA \cite{murray2012ava}, 120,000 from COCO \cite{lin2014microsoft} training  set, 33,000 from VOC \cite{everingham2010pascal}, 400,000 from Places365 \cite{zhou2017places}, and 1,850 from CERTH-Blur \cite{mavridaki2014no}. These datasets were originally built for various other purposes, e.g.,
the AVA dataset is for aesthetic visual analysis, VOC and COCO datasets are introduced for object recognition task. Thus, the labels for these images are also diverse. In brief, the images collected from CERTH-Blur are without any kinds of labels while the remaining images have scene category labels. The total number of category for these images is 26,332. 

The dataset compilation process results in a well-balanced distribution of synthetic and authentic distortion images. Specifically, we acquire 840,000 synthetically distorted images and 809,850 authentically distorted images, yielding an approximate 1:1 ratio between the two categories. This balanced composition enhances the dataset's utility for training and evaluating machine learning models in image distortion analysis.

\textit{Content, Distortion and Appearance Annotation Texts}: After collecting more than 1.6 million images, we need to prepare textual descriptions for them. As the joint contrastive learning is designed for the BIQA, the textual descriptions should be highly related to visual quality of images. It has been demonstrated that the visual quality of an image depends on its content, its distortion, and the interaction between them \cite{10196496}.
The semantic content of an image pertains to its high-level information, encompassing elements such as objects, scenes, and contextual relationships. In contrast, both distortions and appearance are concerned with low-level image quality attributes. Specifically, distortion represents explicit aspects of image quality, whereas appearance embodies implicit characteristics that influence perceptual assessment. Thus, the annotation texts should describe the images content, distortion and their appearance characteristics.

For image content description text, we adopt the textual template format of ``A photo of a(n) $\left \{s \right \} $", where $\left \{s \right \} $ is given by the scene label or object label of the image. This template allows us to summarize the overall content information of the image based on $\left \{s \right \} $. Fortunately, most images collected above contain scene or object labels. For the images without semantic labels, we obtain their content description texts by employing the image captioning method in \cite{mokady2021clipcap}. To ensure the quality and consistency of the caption, we apply length-based filtering to eliminate excessively short or long captions. Subsequently, we conduct a manual review to verify the accuracy and relevance of the captions.

For the images containing synthetic distortions, we annotate them with their distortion labels. We observe that some distortion types give rise to very similar visual appearances, such as Gaussian blur and lens blur. Thus, when preparing textual descriptions of synthetic distortions, due to the similarity of the languages in the visual senses, we combine the 25 distortion types from the KADIS dataset into 19 types (details in Appendix A in the supplementary materials) . Thus, there are a total of 95 kinds of distortion (19 types × 5 degrees). For each of these 19 distortion types, we design a set of 10 possible descriptive phrases for it (see Appendix A in the supplementary materials). For each of the 5 degrees of distortion, we design a list of adjectives, adverbs, and quantifiers for it as well (see Appendix A in the supplementary materials). 
To annotate the images with synthetic distortions, we designed one list of phrases describing distortion types and another list of adjectives, adverbs, and quantifiers describing the degree of distortions.
To annotate an image, we randomly select a phrase from the list corresponding to the distortion type and choose another from the list corresponding to the degree of distortion.

For example, for a given image that contains ``blur" distortion with a degree of 3, the textual description will be one of the phrases randomly picked from the 10 phrases describing ``blur". The adjectives, adverbs or quantifier in that phrase will be randomly picked from the list corresponding to degree 3. The annotation text for this image may be ``some blurring". It is clear that for the same image, there are many equivalent annotations.

For the 140,000 pristine images with high quality, we design another text set consisting of 20 descriptive phrases (see Appendix A in the supplementary materials). These phrases are used to capture possible ways of describing images with high quality. For any one of these images, a phrase randomly picked from these 20 is assigned as its annotation. Note that one image can be assigned multiple annotations.

\begin{figure*}[htbp]
\centering
\subfigure[Brightness]{
\includegraphics[width=0.225\textwidth]{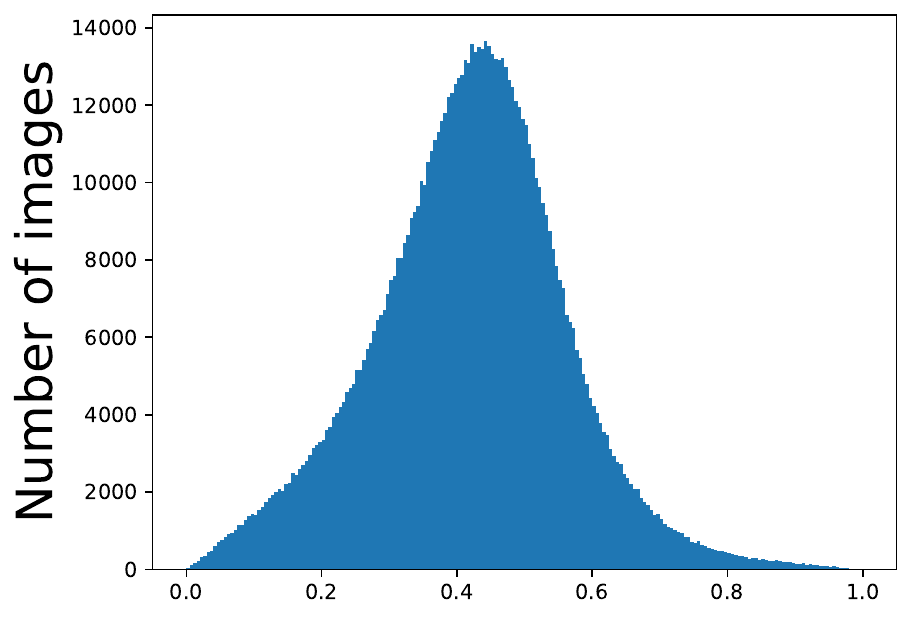}}
\subfigure[Contrast]{
\includegraphics[width=0.225\textwidth]{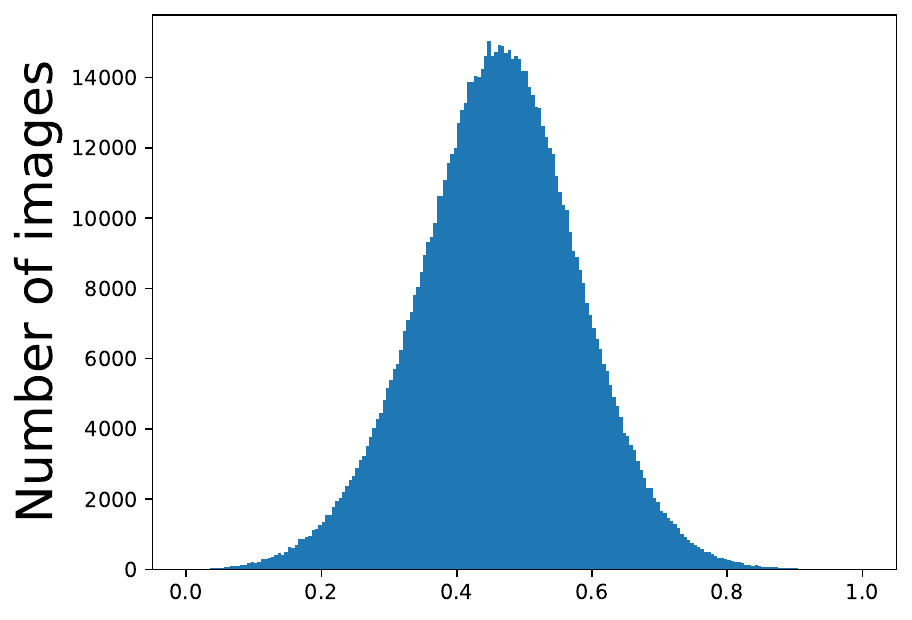}}
\subfigure[Sharpness]{
\includegraphics[width=0.225\textwidth]{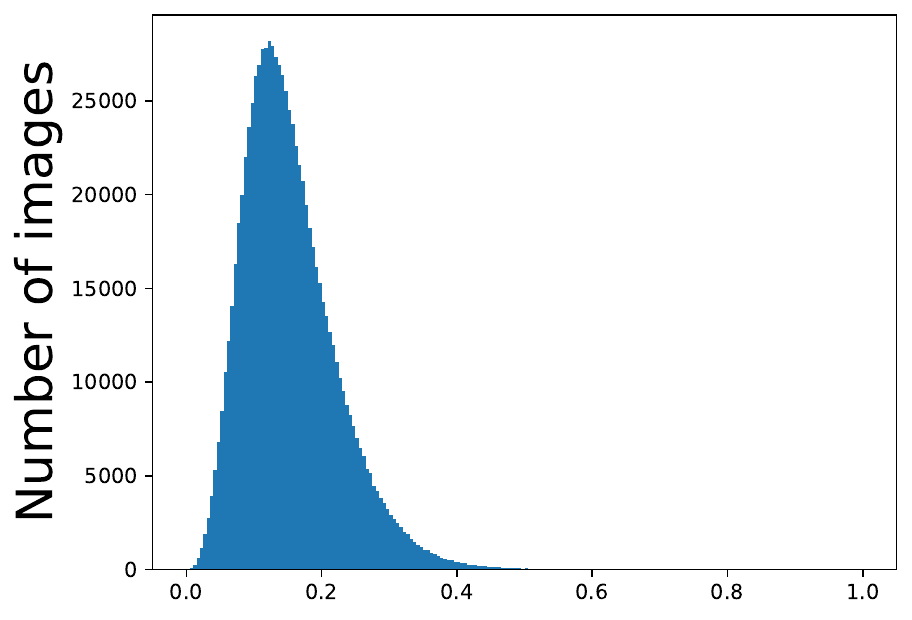}}
\subfigure[Colorfulness]{
\includegraphics[width=0.225\textwidth]{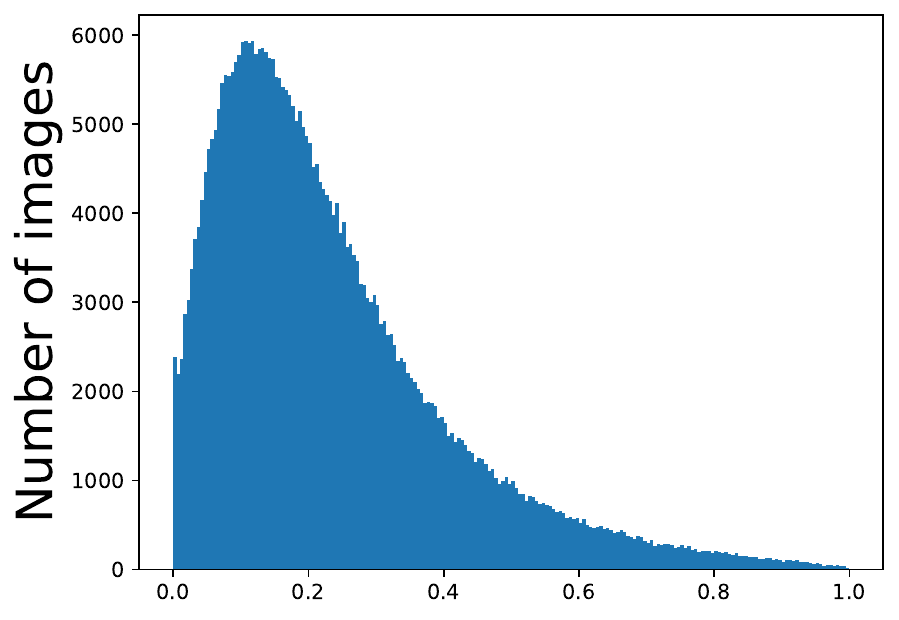}}
\caption{The distribution of four quality relevant aspects of authetic image in TADAC (a)Brightness, (b)Contrast, (c)Sharpness and (d)Colorfulness}.
\label{appearancc_distribution}
\end{figure*}
For images containing authentic (real) distortions, they are much harder to describe textually because they may contain complex and multiple distortions. As the distortions are unknown, our approach is to annotate these images by their appearance characteristics based on 4 quality relevant aspects: brightness, contrast, sharpness, and colorfulness \cite{6202326,10.1117/12.477378}. For a given image, we first calculate its brightness (BR), contrast (CT), sharpness (SH), and colorfulness (CL) (detailed procedures for calculating these are described in Appendix B in the supplementary materials). The distribution of the four appearance attribute values are shown in Fig. \ref{appearancc_distribution}. Analysis revealed a distribution skewed towards median values, with fewer extremes. To balance the dataset, we manually established thresholds for BR, CT, SH, and CL values, categorizing them as ``high," ``medium," or ``low." This manual classification scheme is calibrated to ensure an approximately equal number of images across the three categories.
Recent studies have demonstrated the efficacy of single-modal GPT \cite{gpt4} to be a reliable evaluation instrument for purely linguistic tasks \cite{gptjudging}.

For each level within each visual aspect, we generate a set of descriptive phrases, such as ``$\left \{l \right \}$ contrast," ``The image is $\left \{l \right \}$ contrast" or ``The contrast in image is $\left \{l \right \}$", where $l$ is the level of the contrast. While these phrases may vary lexically, they maintain semantic consistency.
In total we have $4\times3\times20= 240$ phrases for annotating images containing real distortions (see Appendix B in the supplementary material).
For a given image, we label each aspect with a phrase randomly picked from the list corresponding to its value category. For example, for an image whose CT has a ``medium" value, we randomly pick one of the 20 phrases corresponding to medium CT value.
Note that in the current version of the TADAC dataset.
These appearance descriptions were assigned to the subset of images containing real distortions, they can be easily applied to the whole database.

Based on the above procedures, we have constructed a database containing images of diverse contents, a variety of distortions and rich appearances. Importantly, these images are annotated with texts describing their semantic contents, distortion characteristics and appearance properties. These texts can represent quality relevant high level knowledge. When used to train the joint contrastive representation learning framework in Fig. \ref{SLIQUE_Model}, they should enable the system to capture these quality relevant knowledge in its representation vectors. Obviously, the database can be used in other contexts and will be particularly useful for developing IQA applications.   

\subsection{Training Sample Pairing Strategy }
\label{Pairing Strategy}
It is required to produce positive and negative sample pairs for both the vision-language and the visual contrastive learning. Clearly, the pairing strategy should be appropriate for the BIQA task. The optimization objective of image-language branch is to judge whether the input text corresponds to the visual quality of the input image.
Thus, an image-language pair is a positive one if the text comes from the descriptions of the content, distortion or appearance of the image. Otherwise, 
the pair is a negative one.  
For the image-image branch, we follow the same pairing scheme as in \cite{saha2023re}, where two sub-images cropped from an image with 10$\%$ to 30$\%$ overlapping are treated as a positive pair, and two sub-images cropped from different images, exhibiting distinct content or distortion types, are classified into separate categories and thus form a negative pair.

\subsection{Comparisons with Previous BIQA  Works}
As far we are aware, SLIQUE is the first BIQA model that explicitly models all three types visual quality relevant attributes, semantic contents, distortion characteristics and appearance properties.  Our method is joint contrastive learning which combines vision-language supervised contrastive learning and visual self-supervised contrastive learning, and can take full advantage of these two contrastive learning methods. With the joint contrastive learning mechanism, our model is distinguished from the models that only use a vision-language supervised contrast learning mechanism (such as CLIP-IQA \cite{wang2023exploring} and LIQE \cite{zhang2023blind}), as well as the models with only visual self-supervised contrast learning (such as CONTRIQUE and Re-IQA).
Removing either contrastive learning degrades the performance of the model as will be detailed in the ablation study of Table \ref{individual_experiment}. Furthermore, we have constructed the first of a kind database specifically designed for learning all three categories of quality bearing image features. We have developed a systematic approach to constructing quality relevant image textual labels for exploiting vision-language modeling technology to advance research. 

\begin{table*}[t]
\centering
\caption{Performance comparisons of SLIQUE against various BIQA models on IQA databases with synthetic distortions.
 Higher SROCC and PLCC imply better performance. The entry marked as `-' means the results are unavailable.}
\fontsize{9}{11}\selectfont
\begin{tabular}{|c|cccccccc|cc|}
\hline
\multirow{2}{*}{Method}                                                                                                                      & \multicolumn{2}{c|}{LIVE}                               & \multicolumn{2}{c|}{CSIQ}                               & \multicolumn{2}{c|}{TID2013}                            & \multicolumn{2}{c|}{KADID}  &\multicolumn{2}{c|}{Weighted Average}       \\ \cline{2-11} 
                        & \multicolumn{1}{c}{SROCC$\uparrow$} & \multicolumn{1}{c|}{PLCC$\uparrow$}  & \multicolumn{1}{c}{SROCC$\uparrow$} & \multicolumn{1}{c|}{PLCC$\uparrow$}  & \multicolumn{1}{c}{SROCC$\uparrow$} & \multicolumn{1}{c|}{PLCC$\uparrow$}  & \multicolumn{1}{c}{SROCC$\uparrow$} & \multicolumn{1}{c|}{PLCC$\uparrow$} &\multicolumn{1}{c}{SROCC$\uparrow$} & \multicolumn{1}{c|}{PLCC$\uparrow$} \\ \hline

BRISQUE                 & 0.939                      & \multicolumn{1}{c|}{0.935} & 0.746                      & \multicolumn{1}{c|}{0.829} & 0.604                      & \multicolumn{1}{c|}{0.694} & 0.528                      & 0.567 &0.580 	&0.629 
	
\\
NIQE                    & 0.907                      & \multicolumn{1}{c|}{0.901} & 0.627                      & \multicolumn{1}{c|}{0.712} & 0.315                      & \multicolumn{1}{c|}{0.393} & 0.374                      & 0.428 &0.408 	&0.464 
	
\\
CORNIA                  & 0.947                      & \multicolumn{1}{c|}{0.950} & 0.678                      & \multicolumn{1}{c|}{0.776} & 0.678                      & \multicolumn{1}{c|}{0.768} & 0.516                      & 0.558 &0.583 	&0.636 
	
\\
DB-CNN                  & 0.968                      & \multicolumn{1}{c|}{0.971} & 0.946                      & \multicolumn{1}{c|}{0.959} & 0.816                      & \multicolumn{1}{c|}{0.865} & 0.851                      & 0.856 &0.855 	&0.870 
	
\\
WaDIQaM                 & 0.955                      & \multicolumn{1}{c|}{0.960} & 0.852                      & \multicolumn{1}{c|}{0.844} & 0.835                      & \multicolumn{1}{c|}{0.855} & 0.739                      & 0.752 &0.777 	&0.790 
	
\\
PQR                     & 0.965                      & \multicolumn{1}{c|}{0.971} & 0.872                      & \multicolumn{1}{c|}{0.901} & 0.740                      & \multicolumn{1}{c|}{0.798} & -                          & -     &-	&-
	
\\
P2P-BM                  & 0.959                      & \multicolumn{1}{c|}{0.958} & 0.899                      & \multicolumn{1}{c|}{0.902} & 0.862                      & \multicolumn{1}{c|}{0.856} & 0.840                      & 0.849 &0.854 	&0.859 
	
\\
HyperIQA                & 0.962                      & \multicolumn{1}{c|}{0.966} & 0.942                      & \multicolumn{1}{c|}{0.955} & 0.840                      & \multicolumn{1}{c|}{0.858} & 0.852                      & 0.845 &0.860 	&0.860 
	
\\
MetaIQA                 & 0.960                      & \multicolumn{1}{c|}{0.959} & 0.899                      & \multicolumn{1}{c|}{0.902} & 0.856                      & \multicolumn{1}{c|}{0.868} & 0.853                      & 0.846 &0.862 &	0.860 
	
\\
MUSIQ                   & 0.837                      & \multicolumn{1}{c|}{0.818} & 0.697                      & \multicolumn{1}{c|}{0.766} & -                          & \multicolumn{1}{c|}{-}     & 0.572                      & 0.584 &-	&-
	
\\
UNIQUE                  & 0.969                      & \multicolumn{1}{c|}{0.968} & 0.902                      & \multicolumn{1}{c|}{0.927} & -                          & \multicolumn{1}{c|}{-}     & 0.878                      & 0.876 &-	&-
	
\\
CONTRIQUE               & 0.960                      & \multicolumn{1}{c|}{0.961} & 0.947                      & \multicolumn{1}{c|}{0.958} & 0.861                      & \multicolumn{1}{c|}{0.871} & \textbf{0.934}                      & \textbf{0.937} &0.921 	&0.925 
	
\\
Re-IQA                  & 0.970                      & \multicolumn{1}{c|}{0.971} & 0.947                      & \multicolumn{1}{c|}{0.960} & 0.804                      & \multicolumn{1}{c|}{0.861} & 0.872                      & 0.885 &0.867 	&0.889 
	
\\
SAWAR                   & \textbf{0.973}                      & \multicolumn{1}{c|}{\textbf{0.978}} & \textbf{0.961}                      & \multicolumn{1}{c|}{\textbf{0.967}} & \textbf{0.884}                      & \multicolumn{1}{c|}{\textbf{0.896}} & 0.928                      & 0.930  &\textbf{0.923} 	&\textbf{0.927} 
	
\\
CLIP-IQA+               & 0.948                      & \multicolumn{1}{c|}{0.952} & 0.907                      & \multicolumn{1}{c|}{0.928} & 0.835                      & \multicolumn{1}{c|}{0.857} & 0.913                      & 0.909 &0.898 	&0.901 
	
\\ 

GRepQ               & 0.953                      & \multicolumn{1}{c|}{0.958} & 0.941                      & \multicolumn{1}{c|}{0.950} & -                      & \multicolumn{1}{c|}{-} & -                      & - &- 	&- \\
\hline
SLIQUE                  & \textbf{0.982}                      & \multicolumn{1}{c|}{\textbf{0.982}} & \textbf{0.966}                      & \multicolumn{1}{c|}{\textbf{0.973}} & \textbf{0.884}                      & \multicolumn{1}{c|}{\textbf{0.899}} & \textbf{0.957}                      & \textbf{0.959} &\textbf{0.943} 	&\textbf{0.948} 
	
\\ \hline
\end{tabular}
 
 \label{result_syn}
\end{table*}

\section{Experiments}

\subsection{Implementation Details }

The implementation is based on PyTorch and a machine with 8 Nvidia A100-40GB GPUs. In the joint contrastive learning architecture, we employ ResNet-50 followed by an adaptive pooling layer and 2 multilayer perceptron (MLP) layers as the image encoder $F$. The language encoder $G$ is implemented as GPT-2 with 63M parameters \footnote{We are aware there exist newer and more powerful large language models (LLMs). Although a more powerful model could well lead to better results, the goal here is to demonstrate the technical soundness of the designing principle of SLIQUE.}, which uses the byte-pair encoding with a vocabulary size of 49K tokens and a maximum context length of 77. The output features of both the image and the language encoders have a size of 512$\times$1. Most collected images used in the joint contrastive learning have a size of 512$\times$384\footnotemark\footnotetext{Some authentic distortion images are slightly larger than 512$\times$384. We simply resize them to 512$\times$384.}, and all the cropped sub-images (either the saliency based or the OLA based) have a size of 224$\times$224. The collected image data is also augmented in the same manner as \cite{madhusudana2022image}. The temperature parameter in the InfoNCE loss is set to 0.1 by following the suggestion in \cite{oord2018representation}. The default value of $\alpha$ in Equation (4) is 0.7, which will be discussed in Ablation Studies. The SGD optimizer with initial learning rate of 0.006 and batch size of 1024 is used to train the network of joint contrastive learning. Furthermore, the learning rate undergoes a linear warm-up for the first two epochs and then follows a cosine decay schedule without restarts. The training starts from scratch and stops when the epoch number reaches 30.

The ridge regression is learned on some IQA dataset with images labeled with MOS. We include 8 datasets in the experiments, including LIVE \cite{sheikh2006statistical}, CSIQ \cite{larson2010most}, TID2013 \cite{ponomarenko2015image}, and KADID \cite{lin2019kadid}, KonIQ \cite{hosu2020koniq}, CLIVE \cite{ghadiyaram2015massive}, FLIVE \cite{ying2020patches}, and SPAQ \cite{fang2020perceptual}. Among these datasets, the first four are based on synthetic distortions while the last four are built for authentic distortions. The images in SPAQ are resized such that the shorter side is 512 by following the suggestion in \cite{fang2020perceptual}, while the image sizes in other datasets remain unchanged. In each dataset, three sets are randomly splitted for learning, validation, and testing respectively. For the FLIVE dataset, we follow the splitting scheme recommended by its authors. For the other 7 datasets, 70$\%$ of the data are used for learning, 10$\%$ are the validation set, and the remaining are used for testing. As the learning of ridge regression requires no iteration, we use the validation set to determine the regularization coefficient in the regression. The random splitting is repeated 10 times, and the mean results of the testing set are recorded to avoid possible bias. The above protocols of image resizing and dataset splitting have been adopted in many recent researches \cite{madhusudana2022image,10196496,saha2023re}. Thus, we follow the same protocols to make the following comparisons fair. Following the model's training phase, we evaluate SLIQUE performance on a machine equipped with an Nvidia RTX 4090 GPU. For input images with a resolution of 224×224, the inference time of SLIQUE is approximately 1.812 seconds with a FLOPs of 35.04G. The parameter size of model is 76.67M.

\begin{table*}[t]
\centering
 \caption{Performance comparisons of SLIQUE against various BIQA models on IQA databases with authentic distortions.}
\fontsize{9}{11}\selectfont
\begin{tabular}{|c|cccccccc|cc|}
\hline
\multirow{2}{*}{Method} 
                        & \multicolumn{2}{c|}{KonIQ}                              & \multicolumn{2}{c|}{CLIVE}                              & \multicolumn{2}{c|}{FLIVE}                              & \multicolumn{2}{c|}{SPAQ}      &   \multicolumn{2}{c|}{Weighted Average} \\ \cline{2-11} 
                        & \multicolumn{1}{c}{SROCC$\uparrow$} & \multicolumn{1}{c|}{PLCC$\uparrow$}  & \multicolumn{1}{c}{SROCC$\uparrow$} & \multicolumn{1}{c|}{PLCC$\uparrow$}  & \multicolumn{1}{c}{SROCC$\uparrow$} & \multicolumn{1}{c|}{PLCC$\uparrow$}  & \multicolumn{1}{c}{SROCC$\uparrow$} & \multicolumn{1}{c|}{PLCC$\uparrow$}   & \multicolumn{1}{c}{SROCC$\uparrow$} & \multicolumn{1}{c|}{PLCC$\uparrow$}\\ \hline

BRISQUE                 & 0.665                      & \multicolumn{1}{c|}{0.681} & 0.608                      & \multicolumn{1}{c|}{0.629} & 0.288                      & \multicolumn{1}{c|}{0.373} & 0.809                      & 0.817 &0.342 	&0.419 
	
\\
NIQE                    & 0.531                      & \multicolumn{1}{c|}{0.538} & 0.455                      & \multicolumn{1}{c|}{0.483} & 0.211                      & \multicolumn{1}{c|}{0.288} & 0.700                      & 0.709 &0.260 	&0.329 
	
\\
CORNIA                  & 0.780                      & \multicolumn{1}{c|}{0.795} & 0.629                      & \multicolumn{1}{c|}{0.671} & 0.311                      & \multicolumn{1}{c|}{0.356} & 0.709                      & 0.725 &0.363 	&0.405 
	
\\
DB-CNN                  & 0.875                      & \multicolumn{1}{c|}{0.884} & 0.851                      & \multicolumn{1}{c|}{0.869} & 0.435                      & \multicolumn{1}{c|}{0.652} & 0.911                      & 0.915 &0.491 	&0.682 
	
\\
WaDIQaM                 & 0.797                      & \multicolumn{1}{c|}{0.805} & 0.671                      & \multicolumn{1}{c|}{0.680} & 0.571                      & \multicolumn{1}{c|}{0.430} & 0.840                      & 0.845 &0.600 	&0.477 
	
\\
PQR                     & 0.880                      & \multicolumn{1}{c|}{0.884} & 0.857                      & \multicolumn{1}{c|}{0.882} & -                          & \multicolumn{1}{c|}{-}     & -                          & -     &-	&-
	
\\
P2P-BM                  & 0.872                      & \multicolumn{1}{c|}{0.885} & 0.844                      & \multicolumn{1}{c|}{0.842} & 0.535                      & \multicolumn{1}{c|}{0.623} & 0.847                      & 0.830 &0.574 	&0.651 
	
\\
HyperIQA                & 0.906                      & \multicolumn{1}{c|}{0.917} & \textbf{0.859}                      & \multicolumn{1}{c|}{0.882} & 0.554                      & \multicolumn{1}{c|}{0.623} & 0.916                      & 0.919 &0.597 	&0.659 
	
\\
MetaIQA                 & 0.850                      & \multicolumn{1}{c|}{0.887} & 0.802                      & \multicolumn{1}{c|}{0.835} & 0.540                      & \multicolumn{1}{c|}{0.507} & 0.822                      & 0.804 &0.576 &	0.548 
	
\\
MUSIQ                   & \textbf{0.916}                      & \multicolumn{1}{c|}{\textbf{0.928}} & 0.828                      & \multicolumn{1}{c|}{0.785} & \textbf{0.646}                      & \multicolumn{1}{c|}{\textbf{0.739}} & 0.917                      & 0.921 &\textbf{0.678} 	&\textbf{0.760} 
	
\\
UNIQUE                  & 0.896                      & \multicolumn{1}{c|}{0.910} & 0.854                      & \multicolumn{1}{c|}{\textbf{0.890}} & -                          & \multicolumn{1}{c|}{-}     & -                          & -     &-	&-
	
\\
CONTRIQUE               & 0.896                      & \multicolumn{1}{c|}{0.901} & 0.845                      & \multicolumn{1}{c|}{0.857} & 0.580                      & \multicolumn{1}{c|}{0.641} & 0.914                      & 0.919 &0.619 &	0.674 
	
\\
Re-IQA                  & 0.914                      & \multicolumn{1}{c|}{0.923} & 0.840                      & \multicolumn{1}{c|}{0.854} & 0.645                      & \multicolumn{1}{c|}{0.733} & \textbf{0.918}                      & \textbf{0.925} &0.677 	&0.756 
	
\\
SAWAR                   & 0.894                      & \multicolumn{1}{c|}{0.906} & 0.846                      & \multicolumn{1}{c|}{0.857} & 0.544                      & \multicolumn{1}{c|}{0.642} & 0.916                      & 0.919 &0.588 	&0.674 
	
\\
CLIP-IQA+               & 0.895                      & \multicolumn{1}{c|}{0.909} & 0.805                      & \multicolumn{1}{c|}{0.832} & 0.540                      & \multicolumn{1}{c|}{0.566} & 0.864                      & 0.866 &0.581 	&0.605 
	
\\ 
GRepQ               & 0.855                      & \multicolumn{1}{c|}{0.861} & 0.822                      & \multicolumn{1}{c|}{0.836} & -                      & \multicolumn{1}{c|}{-} & -                      & - &- 	&- \\

\hline
SLIQUE                  & \textbf{0.916}                      & \multicolumn{1}{c|}{\textbf{0.929}} &\textbf{0.897}                      & \multicolumn{1}{c|}{\textbf{0.901}} & \textbf{0.720}                      & \multicolumn{1}{c|}{\textbf{0.754}} & \textbf{0.921}                      & \textbf{0.925} &\textbf{0.744} 	&\textbf{0.775} 
	
\\ \hline
\end{tabular}
  
\label{authentic}
\end{table*}
\begin{table*}[t]
\centering
 \caption{Performance comparisons of SLIQUE against various BIQA models on underwater IQA databases.}
\fontsize{9}{11}\selectfont
\begin{tabular}{|c|cccc|cc|}
\hline
\multirow{2}{*}{Method}                                                                                                                 & \multicolumn{2}{c|}{UIF}                               & \multicolumn{2}{c|}{UID2021}                               &         \multicolumn{2}{c|}{Weighted Average}       
\\ \cline{2-7} 
                        & \multicolumn{1}{c}{SROCC$\uparrow$} & \multicolumn{1}{c|}{PLCC$\uparrow$}  & \multicolumn{1}{c}{SROCC$\uparrow$} & \multicolumn{1}{c|}{PLCC$\uparrow$}  & \multicolumn{1}{c}{SROCC$\uparrow$} & \multicolumn{1}{c|}{PLCC$\uparrow$} \\ \hline
HyperIQA                    & 0.290                      & \multicolumn{1}{c|}{0.311}  & 0.476                      & \multicolumn{1}{c|}{0.450}  & 0.383 	&0.381 
\\
MUSIQ                   & 0.309                       & \multicolumn{1}{c|}{0.285}  & 0.403                       & \multicolumn{1}{c|}{0.383}  & 0.281 	    &0.265 
\\
CONTRIQUE                   & \textbf{0.704}                       & \multicolumn{1}{c|}{\textbf{0.731}}  & \textbf{0.833}                       & \multicolumn{1}{c|}{\textbf{0.830}}  & \textbf{0.770} 	    & \textbf{0.781}
\\
CLIP-IQA+                   & 0.425                       & \multicolumn{1}{c|}{0.429}  & 0.298                       & \multicolumn{1}{c|}{0.328}  & 0.251 	    &0.267 
\\
\hline
SLIQUE                  & \textbf{0.766}                      & \multicolumn{1}{c|}{\textbf{0.792}} & \textbf{0.860}          & \multicolumn{1}{c|}{\textbf{0.853}} & \textbf{0.814} 	&\textbf{0.823}
\\
\hline
\end{tabular}
 
 \label{result_underwater}
\end{table*}

\begin{table*}[t]
\centering
 \caption{Performance comparisons of SLIQUE against various BIQA models on enhanced IQA databases.}
\fontsize{9}{11}\selectfont

\begin{tabular}{|c|cccc|cc|}
\hline
\multirow{2}{*}{Method}                                                                                                                 & \multicolumn{2}{c|}{QADS}                               & \multicolumn{2}{c|}{GFIQA}                               &         \multicolumn{2}{c|}{Average}       
\\ \cline{2-7} 
                        & \multicolumn{1}{c}{SROCC$\uparrow$} & \multicolumn{1}{c|}{PLCC$\uparrow$}  & \multicolumn{1}{c}{SROCC$\uparrow$} & \multicolumn{1}{c|}{PLCC$\uparrow$}  & \multicolumn{1}{c}{SROCC$\uparrow$} & \multicolumn{1}{c|}{PLCC$\uparrow$} \\ \hline
HyperIQA                    & 0.747                      & \multicolumn{1}{c|}{0.710}  & 0.880                  & \multicolumn{1}{c|}{0.892}  & 0.814	&0.801 
\\
MUSIQ                    & 0.772                      & \multicolumn{1}{c|}{0.768}  & 0.910                      & \multicolumn{1}{c|}{0.901}  & 0.841 	&0.835 
\\
CONTRIQUE                   & \textbf{0.935}                       & \multicolumn{1}{c|}{\textbf{0.938}}  & \textbf{0.911}                      & \multicolumn{1}{c|}{\textbf{0.914}}  & \textbf{0.923} 	    & \textbf{0.926} 
\\
CLIP-IQA+                   & 0.675                       & \multicolumn{1}{c|}{0.680}  & 0.870                       & \multicolumn{1}{c|}{0.877}  & 0.773 	    &0.779 
\\
\hline
SLIQUE                  & \textbf{0.959}                      & \multicolumn{1}{c|}{\textbf{0.953}} & \textbf{0.932}          & \multicolumn{1}{c|}{\textbf{0.930}} & \textbf{0.946} 	&\textbf{0.942}
\\
\hline
\end{tabular}
 
 \label{result_enhanced}
\end{table*}
\subsection{Results and Comparisons}
 We test our model on the above mentioned 8 IQA datasets and compare it with 15 BIQA models. The competitors include 16  BIQA methods: BRISQUE \cite{mittal2012no},  NIQE\cite{mittal2012making}, CORNIA \cite{ye2012unsupervised}, WaDIQaM \cite{bosse2017deep}, PQR \cite{zeng2017probabilistic}, DB-CNN\cite{zhang2018blind}, P2P-BM\cite{ying2020patches}, HyperIQA\cite{su2020blindly}, MetaIQA\cite{zhu2020metaiqa}, MUSIQ\cite{ke2021musiq}, UNIQUE \cite{zhang2021uncertainty}, CONTRIQUE\cite{madhusudana2022image}, Re-IQA\cite{saha2023re}, SAWAR
 \cite{10196496}, CLIP-IQA\cite{wang2023exploring}, and GRepQ\cite{srinath2024learning}.  Additionally, we evaluate our model on other types of IQA datasets and compare it with models including HyperIQA\cite{su2020blindly}, MUSIQ\cite{ke2021musiq}, CONTRIQUE\cite{madhusudana2022image}, CLIP-IQA\cite{wang2023exploring} on underwater images UIF\cite{uif}, UID2021\cite{uid2021}, super resolution images QADS\cite{QADS} and human facial images GFIQA\cite{faceiqa}. All results of BIQA are either from the original papers or reproduced by their source codes.

Two widely-used criteria Spearman rank-order correlation coefficient (SROCC) and the Pearson linear correlation coefficient (PLCC) are used to measure the performance of different BIQA models. 
Table \ref{result_syn} and Table \ref{authentic} list the quantitative results on synthetic distortion datasets and authentic distortion datasets, respectively. The best and second-best performances are highlighted in bold. 
The proposed method achieves 16 best results across all datasets. It indicates that our model exhibits excellent performances on both synthetic and authentic distortions. We attribute it to the data preparation of both synthetic distortion and authentic images for the joint contrast learning. In contrast, only synthetic distortion images can be employed to train the auto-encoders of SAWAR in a self-supervised way. Thus, the performance of SAWAR on authentic distortion datasets is inferior, although its performance on synthetic distortion datasets is impressive. In the CONTRIQUE and Re-IQA models, both types of distortions can be employed to train the quality-relevant feature extractors. However, they only consider visual contrast learning and ignore the benefits of language supervised learning. In addition, the CLIP-IQA model is mainly devoted to exploit the knowledge already contained in the pre-trained CLIP and finetune with simple statements.
Different from them, the proposed method based on vision-language and visual contrastive learning can achieve superior results by using a more comprehensive text description.
Table \ref{result_underwater} and Table \ref{result_enhanced} list the quantitative results on underwater and enhanced images datasets, respectively. The best and second-best results are highlighted in bold. HyperIQA, MUSIQ, CLIP-IQA exhibit significant performance degradation on the underwater dataset. This degradation may be due to the unique distortions in underwater images, which differ significantly from those in standard image quality assessments. The difference in performance is less in QADS, while the discrepancy in performance is least in GFIQA, which can be attributed to its sole composition of human facial images, thus presenting a limit range of content variability. This consistent outperformance on different types of IQA datasets underscores the robustness of SLIQUE and highlights the generalization of the representations learned by the model.

\begin{figure*}[t]
    \centering
    \subfigure[CONTRIQUE]{
    \includegraphics[width=0.45\textwidth]{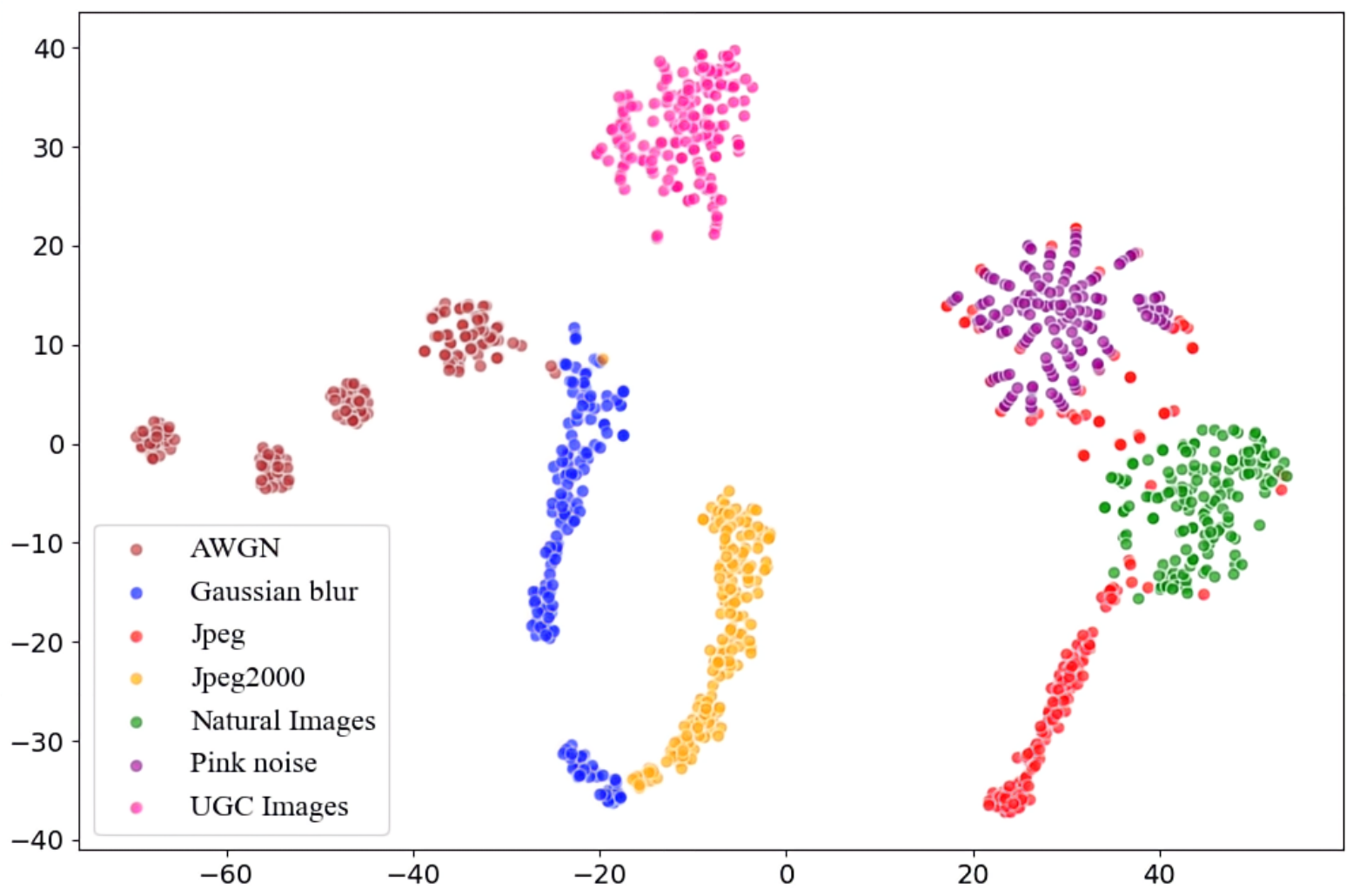}}
    \hspace{0.01\linewidth}
    \subfigure[SLIQUE]{
    \includegraphics[width=0.45\textwidth]{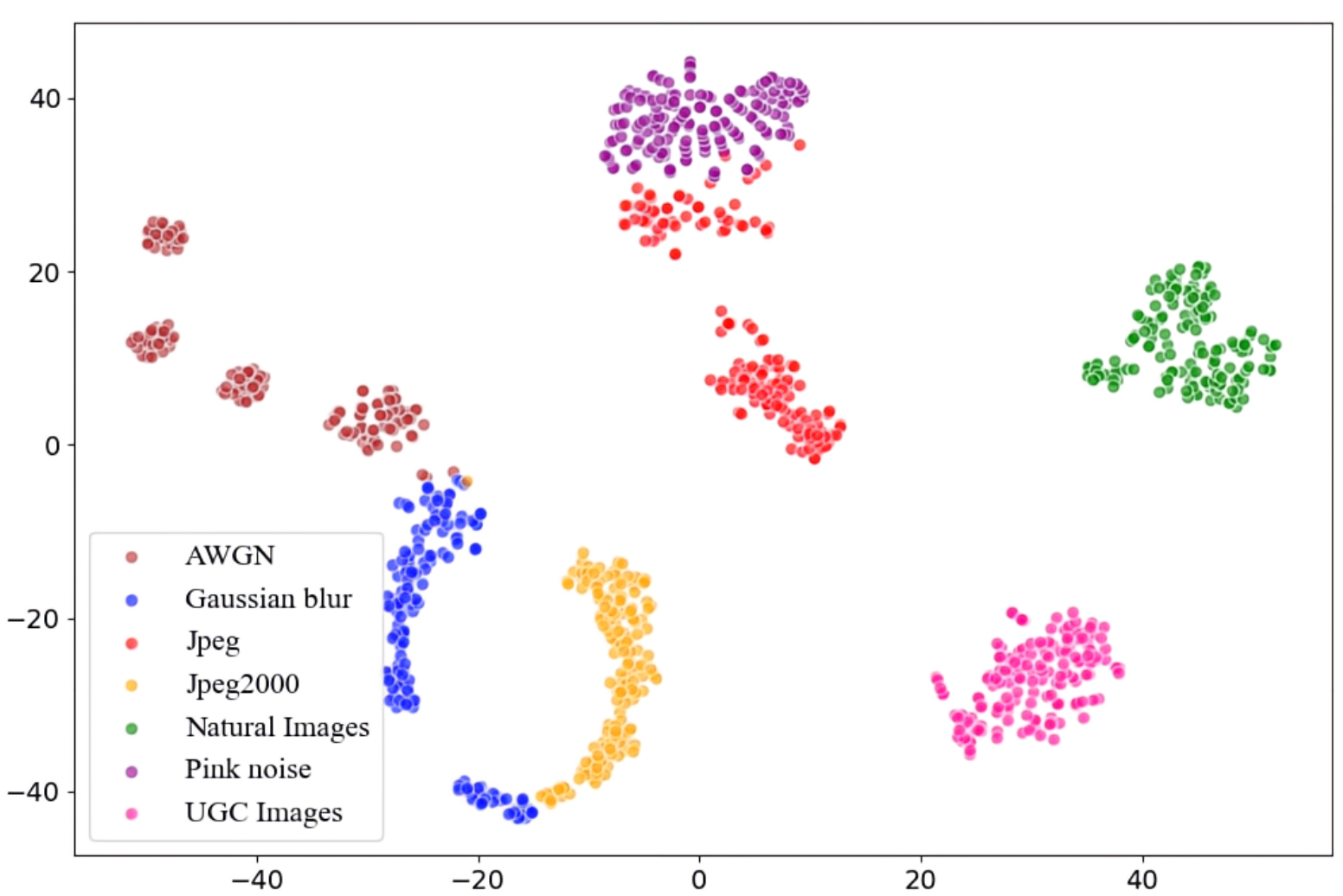}}
    \vfill
    \centering
    \subfigure[CONTRIQUE]{
    \includegraphics[width=0.45\textwidth]{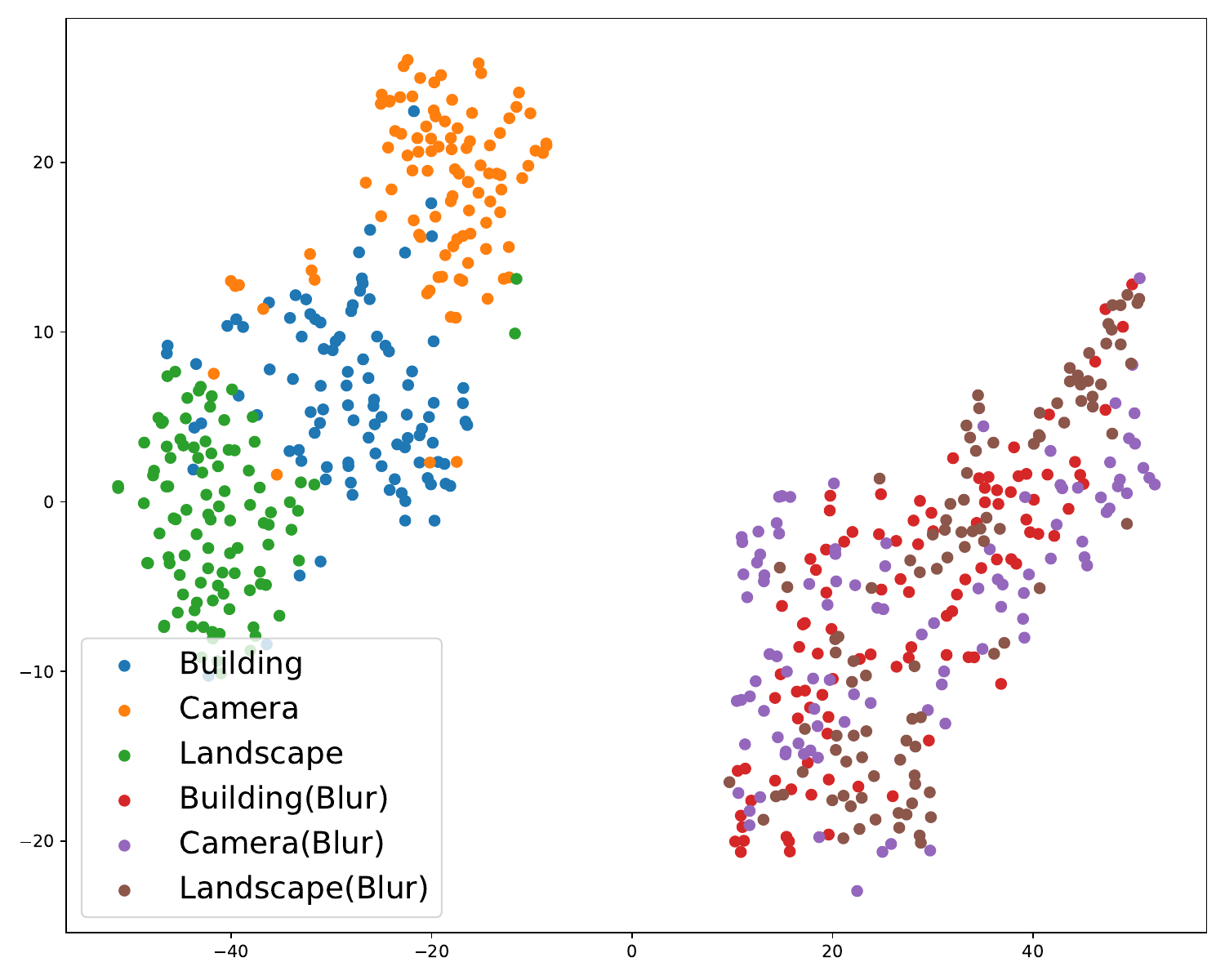}}
        \hspace{0.01\linewidth}
    \subfigure[SLIQUE]{
    \includegraphics[width=0.45\textwidth]{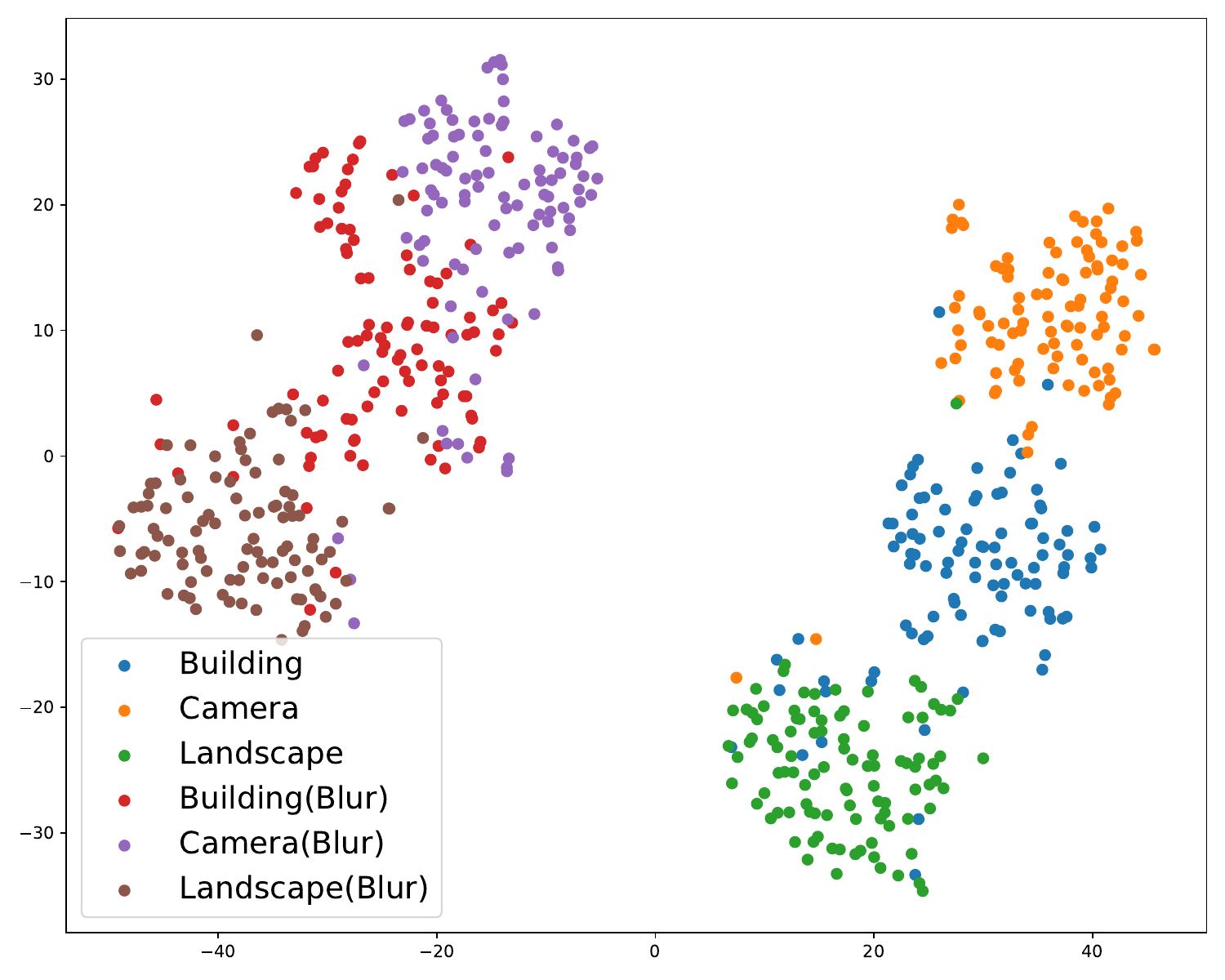}}
\caption{2D t-SNE visualizations of learned representations. For both CONTRIQUE and SLIQUE, we conducted 2D t-SNE visualization experiments using 7 different types of distorted images from 3 image databases and 3 different types of images from KADIS database.}
\label{t_SNE_main}
\end{figure*}

\begin{figure}[h]
\centering \selectfont 
\includegraphics[width=0.45\textwidth]{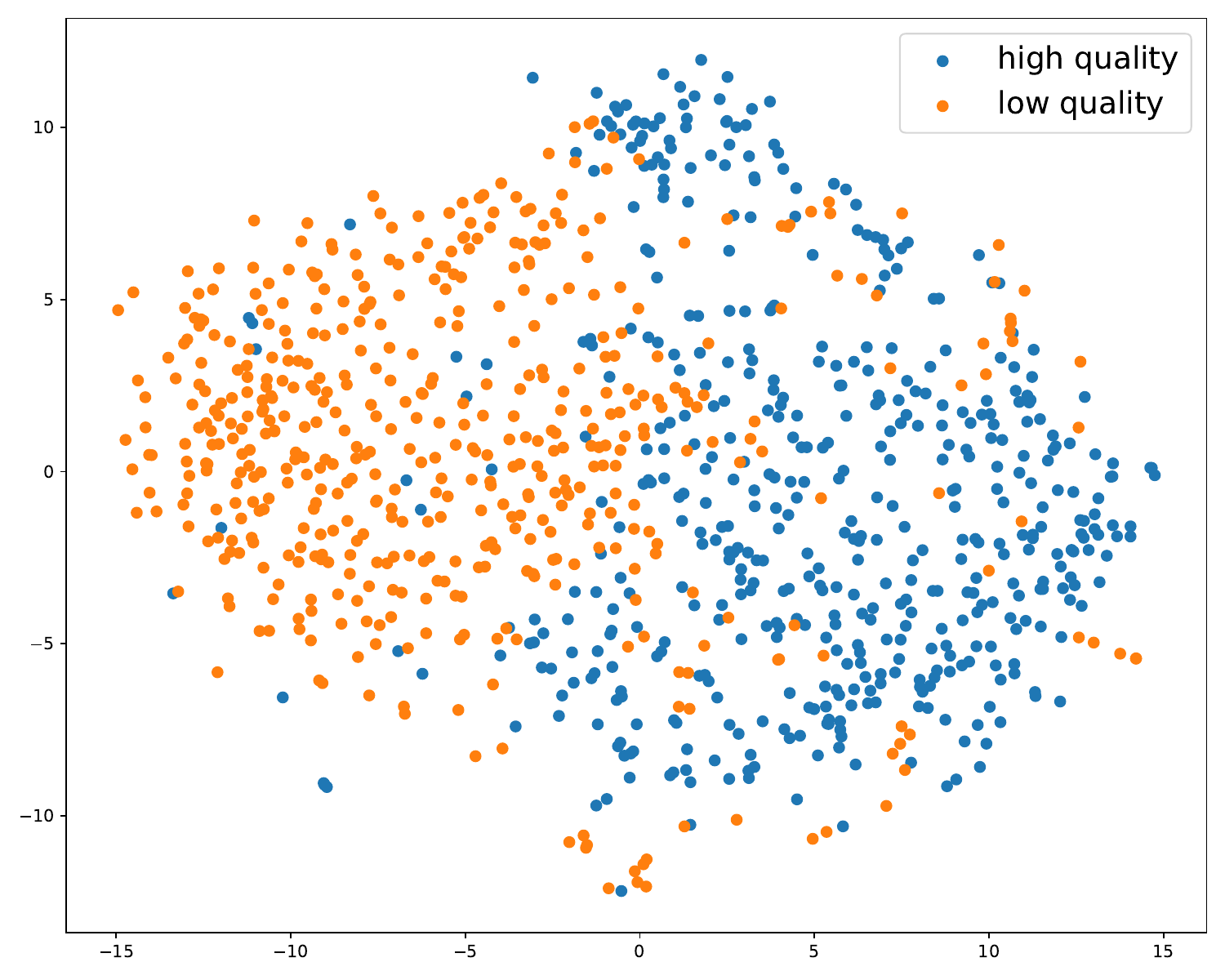}
\caption{2D t-SNE visualizations of learned representations for 1,000 images in the Wild from KonIQ.}
\label{t_SNE_sub}
\end{figure}

\subsection{T-SNE Visualization of Quality Relevant Features}
The success of SLIQUE relies on its image encoder's ability to extract discriminative image features that can capture information related to factors affecting image quality such as semantic content, distortion characteristics and appearance properties. Fig. \ref{t_SNE_main} shows examples of t-SNE visualization of the learned representations of SLIQUE and that of a recent method CONTRIQUE \cite{madhusudana2022image}. 
We used 750 images sampled from CSIQ (Synthetic distortions - \#600) and User-Generated Content (UGC) images from KonIQ (UGC -\#150) and pristine images from KADIS (Natural images - \#150). It is clearly seen that, for the CONTRIQUE model, some features from differently distorted images are not well distinguished from each other, e.g., the JPEG and the Pink noise. 
In contrast, the representations learned by our new SLIQUE can better separate different distortion types which may explain why it has superior performances.
Fig. \ref{t_SNE_main} illustrates additional t-SNE visualizations of 300 pristine images from KADIS (100 each of buildings, cameras, and landscapes), distorted by Gaussian blur. The learned representations of CONTRIQUE model exhibit limited discriminative capacity across images with identical distortion types but varying content. In contrast, the learned representations of SLIQUE  demonstrate more pronounced inter-class separation. Fig. \ref{t_SNE_sub} shows t-SNE visualizations of 1,000 KonIQ images, split evenly between low (\textless33) and high (\textgreater67) quality scores. The plot exhibits distinct clustering, with high-quality images (blue) and low-quality images (orange) forming separate groups. This pattern demonstrates the ability of SLIQUE to learn representations that effectively differentiate image quality. The overlapping region of the clusters represents instances where the predictions may diverge from ground truth. 
These results highlight the strong discriminative power of in distinguishing both content and quality variations, suggesting its effectiveness for image quality assessment tasks.
\subsection{Cross Dataset Evaluations}
To show the generalization ability of SLIQUE, we conduct cross dataset evaluations by training and testing on different datasets. Specifically, the ridge regression of SLIQUE is learned by all the data from a dataset while the parameters of image encoder $F$ are still frozen. Then, the model is tested on all the data from another dataset. Five datasets, including three synthetic and two authentic distortion datasets, are involved in the cross dataset evaluation. The results based on SROCC are presented in Table \ref{cross_dataset}, where the performances of two recently-developed models are also listed. As shown in Table \ref{cross_dataset}. The best and second-best performances are highlighted in bold. Comparing the results in Table II and Table V, we can see that cross-dataset evaluations are more challenging. It is mainly attributed to the difference of the data distribution between two IQA datasets, especially the difference between synthetic and authentic distortion datasets.
Nevertheless, our model outperforms the Re-IQA and the SAWAR, which also take advantage of unsupervised learning to obtain quality-relevant features. It demonstrates that our joint contrastive learning is highly effective in representing the image quality.
\begin{table}[h]
\centering
\caption{Cross database evaluations of IQA models.}
\fontsize{9}{11}\selectfont
\begin{tabular}{|c|c||c|c|c|}
\hline
Training & Testing  &Re-IQA&SAWAR&SLIQUE    \\ \hline
LIVE    & CSIQ      &0.808  &\textbf{0.898}  & \textbf{0.903} \\ 
CSIQ    & LIVE      &0.929  &\textbf{0.930}  & \textbf{0.934} \\ 
CLIVE   & KonIQ     &\textbf{0.769}  &0.665  & \textbf{0.775} \\ 
KonIQ   & CLIVE     &\textbf{0.794} &0.700  & \textbf{0.832} \\
CLIVE   & CSIQ      &\textbf{0.680}   &0.656   &\textbf{0.721} \\
CSIQ   & CLIVE     &0.446   &\textbf{0.452}   &\textbf{0.527} \\
TID2013   & CLIVE     &\textbf{0.423}   &0.411   &\textbf{0.562} \\
CLIVE   & TID2013      &0.436   &\textbf{0.452}   &\textbf{0.531}    \\
\hline 

\end{tabular}
\label{cross_dataset}
\end{table}

\subsection{Ablation Studies}
Four datasets, i.e., CSIQ, KADID, KonIQ and CLIVE, are used in the ablation studies to investigate the  effectiveness of some important designs and settings in SLIQUE. We have conducted 6 ablation studies, including the discussions on individual contrastive learning, the value of parameter $\alpha$, the starting point of optimization, annotation texts, different sizes of training data, and the usage of language feature.

\textbf{Individual contrastive learning.} We first conducted an ablation study to validate the importance of joint contrastive learning by disabling one of the branches. When disabling the image-image branch, only image-language pairs are used to train the encoders $G$ and $F$ from scratch. Similarly, if the image-language branch is disabled, only image-image pairs are employed to train the image encoder $F$. The results based on SROCC are given in Table \ref{individual_experiment}, which shows a performance reduction if we disable either branch. Moreover, we can see from the results that the visual contrastive learning is greatly important, and the vision-language contrastive learning also provides essential benefits.

\begin{table}[h]
\centering
\caption{Performance of individual contrastive learning.}
\fontsize{9}{10.5}\selectfont
\begin{tabular}{|l||c|c|c|c|}

\hline
\multicolumn{1}{|c||}{Ablation cases}    & CSIQ       & KADID  & KonIQ          & CLIVE                   \\ \hline
 w/o Image-Image  & 0.877          & 0.849 & 0.787          & 0.673           \\ 
 w/o Image-Language       & 0.949          & 0.933 & 0.897          & 0.847                   \\ \hline
 SLIQUE              & \textbf{0.966} & \textbf{0.957} & \textbf{0.916} & \textbf{0.897}  \\ \hline
\end{tabular}

\label{individual_experiment}
\end{table}

\begin{figure}[h]

\centering \fontsize{9}{12}\selectfont

\includegraphics[width=0.5\textwidth]{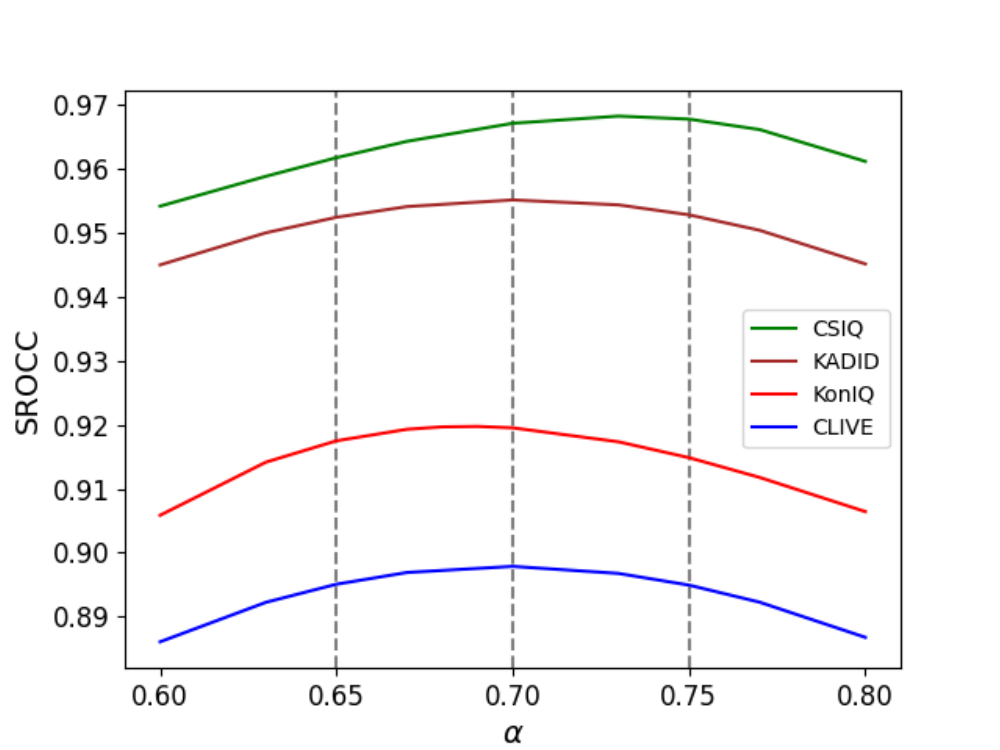}

\caption{Impact of parameter $\alpha$ on the model performance.}
\label{Fig_alpha}
\end{figure}

\textbf{Value of parameter $\alpha$.} 
The parameter $\alpha$ is used to balance the contributions of the image-image branch and the image-language branch. We set a range of $\alpha$ values from 0.6 to 0.8 with a step size of 0.01 to test SLIQUE. The SROCC-based performance is shown in Fig. \ref{Fig_alpha}, where we can see that the model performance is stable and can reach the maximum within the interval of [0.65, 0.75]. Thus, it is reasonable to set the default value of $\alpha$ to 0.7. This value further demonstrates the importance of visual learning as well as the necessity of vision-language learning.

\textbf{Starting point of optimization.} In the joint training of image-language branch and image-image branch, we optimize the language encoder $G$ and the image encoder $F$ from scratch. An alternative strategy is to optimize them (or one of them) from pre-trained models. Here we employed two well-known pre-trained models, i.e., CLIP and SIMCLR, as the starting point of optimization. The results on synthetic distortion datasets and authentic distortion datasets are recorded in Table \ref{pretrained_syn} and Table \ref{pretrained_ugc}, respectively. 
In the first row of each table, the language encoder $G$ is initialized by the pretrained CLIP weights, and the image encoder $F$ is initialized by the pretrained  SIMCLR weights. Other cases are included in the following rows. As the SIMCLR only contains image encoder, it can not used to initialize $G$. From the results in Table \ref{pretrained_syn} and Table \ref{pretrained_ugc}, we find that the optimizations from pretrained models do not result in the best performances, although some performances from them are comparable with state of the art. It may be because that they are originally trained in object recognition or image classification tasks. Thus, using pretrained models is not optimal for the image quality assessment task.

\begin{table}[h]
\centering
\caption{Optimization from pretrained models versus from scratch (synthetic
distortion datasets).}
\fontsize{9}{11}\selectfont
\begin{tabular}{|cc||c|c|}
\hline
\multicolumn{1}{|c|}{Language encoder} &Image encoder & CSIQ          & KADID \\ \hline
\multicolumn{1}{|c|}{from CLIP} &  from SIMCLR & 0.886          & 0.875         \\ \hline
\multicolumn{1}{|c|}{from CLIP} &  from CLIP & 0.912          & 0.907          \\ \hline
\multicolumn{1}{|c|}{from CLIP } &  from scratch & 0.932          & 0.928          \\ \hline
\multicolumn{1}{|c|}{from scratch} &  from CLIP & 0.926          & 0.918          \\ \hline
\multicolumn{1}{|c|}{from scratch} &  from SIMCLR & 0.945          & 0.936         \\ \hline

\multicolumn{2}{|c||}{SLIQUE (both from scratch)}                                            &\textbf{0.966} & \textbf{0.957} \\ \hline
\end{tabular}

\label{pretrained_syn}
\end{table}

\begin{table}[h]
\centering
\caption{Optimization from pretrained models versus from 
scratch (authentic distortion datasets).}
\fontsize{9}{11}\selectfont
\begin{tabular}{|cc||c|c|}
\hline
\multicolumn{1}{|c|}{Language encoder} &Image encoder & KonIQ          & CLIVE \\ \hline
\multicolumn{1}{|c|}{from CLIP} &  from SIMCLR & 0.866          & 0.839         \\ \hline
\multicolumn{1}{|c|}{from CLIP} &  from CLIP & 0.809          & 0.741          \\ \hline
\multicolumn{1}{|c|}{from CLIP} &  from scratch & 0.909          & 0.879          \\ \hline
\multicolumn{1}{|c|}{from scratch} &  from CLIP & 0.884          & 0.765          \\ \hline
\multicolumn{1}{|c|}{from scratch} & from SIMCLR & 0.878          & 0.838         \\ \hline

\multicolumn{2}{|c||}{SLIQUE (both from scratch)}                                            & \textbf{0.916} & \textbf{0.897} \\ \hline
\end{tabular}

\label{pretrained_ugc}
\end{table}

\textbf{Components in annotation texts.} 
To investigate the impact of different components in the annotation texts on quality prediction, we conducted ablation studies by excluding some components in the textual descriptions, resulting in 4 cases. The results based on SROCC are shown in Table \ref{annotation}. In the first row of Table \ref{annotation}, we only keep the content annotation texts, and exclude both the distortion texts for synthetic distortion images and the appearance texts for authentic distortion images. 
The results in the first, second, and third rows respectively indicate that using only content text, distortion text, or appearance attribute text alone is insufficient. In the fourth row, we add the distortion information of synthetic distortion images to the textual descriptions. The results show that the performance on authentic distortion datasets KonIQ and CLIVE even decreases, although the performance on synthetic distortion datasets is improved. It may be attributed to the large differences between the synthetic distortion and the authentic distortion. Similarly, the results of adding the appearance texts are provided in the fifth row which is worse than the full model.
In the sixth row, we report the results of excluding content texts, demonstrating that the content texts are also beneficial to the IQA task.
The comparative analysis conducted above demonstrated the complementary relationship among all three annotation components, thus confirming their effectiveness and essential role in the overall process.
\begin{table}[h]
\centering 
\caption{Ablation studies on components in annotation texts. In each column, the top performance is boldfaced.}
\fontsize{9}{10.5}\selectfont
\begin{tabular}{|l||c|c|c|c|}
\hline
 Annotation texts& CSIQ & KADID & KonIQ & CLIVE  \\ \hline

Content                 & 0.944 & 0.942& 0.901 & 0.884 \\
Distortion             & 0.948 & 0.944 & 0.885 & 0.872 \\
Appearance             & 0.931 & 0.938 & 0.904 & 0.888 \\ 
Content+Distortion      & 0.962 & 0.952& 0.891 & 0.879 \\
Content+Appearance      & 0.950 & 0.946& 0.914 & 0.894  \\
Distort.+Appearance   & 0.963 & 0.953& 0.909 & 0.890 \\ \hline
\multicolumn{1}{|l||}{Full Texts (SLIQUE) }& \textbf{0.966} & \textbf{0.957}& \textbf{0.916} & \textbf{0.897}  \\ \hline
\end{tabular}

\label{annotation}
\end{table}

\textbf{The amount of training.} To investigate the impact of the amount of training data, we divided the database into equal proportions of 20$\%$, 40$\%$, 60$\%$, 80$\%$, and 90$\%$. We can observe from the results in Table \ref{proportion} that using a large number of data is essential in training our two branches. Besides, the results reveal that our method can achieve state-of-the-art performance by using 90$\%$ of the data. 

\begin{table}[h]
\centering 
\caption{SROCC performances of our model for different amount of training data. }
\fontsize{9}{11}\selectfont
\begin{tabular}{|c||c|c|c|c|}
\hline
Datasets proportion & CSIQ   &KADID & KonIQ & CLIVE      \\ \hline
20\%                    & 0.909  &   0.910 & 0.875 & 0.813      \\ 
40\%                    & 0.932    &     0.935 & 0.889 & 0.834      \\ 
60\%                    & 0.946    &     0.941& 0.903 & 0.866      \\ 
80\%                    & 0.953     &     0.949& 0.913 & 0.888     \\ 
90\%                  & \textbf{0.966}  & \textbf{0.957}    & \textbf{0.917} & \textbf{0.897}         \\ \hline
\end{tabular}

\label{proportion}
\end{table}

\textbf{Usage of language feature.} In our SLIQUE, only the image feature is employed in the prediction of visual quality of images. An alternative strategy is to use an image captioning method to produce a textual description for the image to be accessed. With such a strategy, the feature from language encoder can also be fed to the ridge regression. Here we employed the captioning method in \cite{mokady2021clipcap} to produce annotation texts. The results on synthetic distortion datasets and authentic distortion datasets are recorded in Table \ref{features_syn} and Table \ref{features_ugc}, respectively. From the results, we can see that using the language feature is not helpful. It is mainly because the texts produced from the captioning method contain little distortion and appearance information. Furthermore, even if the captioning method can produce perfect texts for the task, the language feature would be redundant due to its similarity to the corresponding image feature. 
\begin{table*}[t]
\centering
\caption{Performance comparisons of SLIQUE and TCL retrained on Q-Pathway and TADAC on IQA databases with synthetic distortions.}
\fontsize{8.8}{11}\selectfont
\begin{tabular}{|c|cccccccc|cc|}
\hline
\multirow{2}{*}{Model} 
                        & \multicolumn{2}{c|}{LIVE}                  & \multicolumn{2}{c|}{CSIQ}                  & \multicolumn{2}{c|}{TID2013}                  & \multicolumn{2}{c|}{KADID}
                        & \multicolumn{2}{c|}{Weighted Average} 
                        \\ \cline{2-11} 
                        &\multicolumn{1}{c}{SROCC$\uparrow$} & \multicolumn{1}{c|}{PLCC$\uparrow$} & \multicolumn{1}{c}{SROCC$\uparrow$} & \multicolumn{1}{c|}{PLCC$\uparrow$} & \multicolumn{1}{c}{SROCC$\uparrow$} & \multicolumn{1}{c|}{PLCC$\uparrow$} & \multicolumn{1}{c}{SROCC$\uparrow$} & \multicolumn{1}{c|}{PLCC$\uparrow$}   & \multicolumn{1}{c}{SROCC$\uparrow$} & \multicolumn{1}{c|}{PLCC$\uparrow$}\\ \hline

TCL(Q-Pathway)               & 0.953           & \multicolumn{1}{c|}{0.942} & 0.946                      & \multicolumn{1}{c|}{0.924} & 0.784                      & \multicolumn{1}{c|}{0.778} & 0.869                      & 0.839 &0.862 	&0.838 

\\
TCL(TADAC)               & 0.964           & \multicolumn{1}{c|}{0.959} & \textbf{0.958}                      & \multicolumn{1}{c|}{0.951} & \textbf{0.863}                      & \multicolumn{1}{c|}{0.859} & \textbf{0.921}                      & \textbf{0.911} & \textbf{0.914} 	& \textbf{0.905 }

\\
SLIQUE(Q-Pathway)                & \textbf{0.971}    & \multicolumn{1}{c|}{\textbf{0.972}} & 0.948                      & \multicolumn{1}{c|}{\textbf{0.957}} & 0.839                      & \multicolumn{1}{c|}{\textbf{0.886}} & 0.892                      & 0.885 &0.889 	&0.895 	
\\
SLIQUE(TADAC)                & \textbf{0.982}    & \multicolumn{1}{c|}{\textbf{0.982}} & \textbf{0.966}                      & \multicolumn{1}{c|}{\textbf{0.973}} & \textbf{0.884}                      & \multicolumn{1}{c|}{\textbf{0.899}} & \textbf{0.957}                      & \textbf{0.959} & \textbf{0.943} 	&\textbf{0.948}
\\
\hline
\end{tabular}
\label{qpathway_syn}
\end{table*}

\begin{table*}[t]
\centering
\caption{Performance comparisons of SLIQUE and TCL retrained on Q-Pathway and TADAC on IQA databases with authentic distortions.}
\fontsize{8.8}{11}\selectfont
\begin{tabular}{|c|cccccccc|cc|}
\hline
\multirow{2}{*}{Model} 
                        & \multicolumn{2}{c|}{KonIQ}                  & \multicolumn{2}{c|}{CLIVE}                  & \multicolumn{2}{c|}{FLIVE}                  & \multicolumn{2}{c|}{SPAQ}
                        & \multicolumn{2}{c|}{Weighted Average} 
                        \\ \cline{2-11} 
                        &\multicolumn{1}{c}{SROCC$\uparrow$} & \multicolumn{1}{c|}{PLCC$\uparrow$} & \multicolumn{1}{c}{SROCC$\uparrow$} & \multicolumn{1}{c|}{PLCC$\uparrow$} & \multicolumn{1}{c}{SROCC$\uparrow$} & \multicolumn{1}{c|}{PLCC$\uparrow$} & \multicolumn{1}{c}{SROCC$\uparrow$} & \multicolumn{1}{c|}{PLCC$\uparrow$}   & \multicolumn{1}{c}{SROCC$\uparrow$} & \multicolumn{1}{c|}{PLCC$\uparrow$}\\ \hline

TCL(Q-Pathway)               & 0.672           & \multicolumn{1}{c|}{0.707} & 0.674                      & \multicolumn{1}{c|}{0.706} & 0.503                      & \multicolumn{1}{c|}{0.533} & 0.834                      & 0.829 &0.526 	&0.556 

\\
TCL(TADAC)               & 0.811           & \multicolumn{1}{c|}{0.831} & 0.708                      & \multicolumn{1}{c|}{0.749} & 0.563                      & \multicolumn{1}{c|}{0.589} & 0.858                      & 0.852 &0.593 	&0.618 

\\
SLIQUE(Q-Pathway)                & \textbf{0.891}    & \multicolumn{1}{c|}{\textbf{0.906}} & \textbf{0.795}                      & \multicolumn{1}{c|}{\textbf{0.819}} & \textbf{0.674}                      & \multicolumn{1}{c|}{\textbf{0.685}} & \textbf{0.916}                      & \textbf{0.920} & \textbf{0.723} 	& \textbf{0.739} 	
\\
SLIQUE(TADAC)              & \textbf{0.916}    & \multicolumn{1}{c|}{\textbf{0.929}} & \textbf{0.897}                      & \multicolumn{1}{c|}{\textbf{0.901}} & \textbf{0.720}                     & \multicolumn{1}{c|}{\textbf{0.754}} & \textbf{0.921}                      & \textbf{0.925} & \textbf{0.744} 	& \textbf{0.774} 
\\
\hline
\end{tabular}
\label{qpathway_ugc}
\end{table*}

\textbf{Benefits of TADAC dataset.} To show the benefits of our TADAC dataset, we would like to re-train the text encoder $G$ and the image encoder $F$ by using another image-text dataset, known as Q-pathway \cite{qins}. Then, the regression is learned on a traditional IQA dataset. Finally, we record the model performances on the traditional dataset. Here, the 8 traditional datasets involved in Section IV.B are tested. The results based on synthetically distorted images are reported in the third row of Table VI, while the results on authentic distortion datasets are given in the third row of Table VII. By comparing the results in the last two rows of both tables, we can see that the performances based on the TADAC are better that those based on the Q-pathway. A probable reason is that the human feedbacks in the Q-pathway might be noisy due to different subjective scales of individuals with various cultural and educational backgrounds. In both tables, we further include the results from a competitive IQA model, known as TCL \cite{yang2022vision}, which also uses image-text pairs to train an image encoder. We train the TCL encoder by using the TADAC and the Q-pathway, respectively. The performances of TCL are provided in the first two rows of Tables VI and VII, which also demonstrate the benefits of our TADAC dataset. Besides, as both results in the first and the third rows are based on the Q-pathway, we can conclude that the SLIQUE outperforms the TCL. The same conclusion can be drawn by comparing the results in the second and the last rows.  
\begin{table}[h]
\centering
\caption{Exploring more features to use in evaluation of synthetic distortion datasets.}
\fontsize{9}{11}\selectfont
\begin{tabular}{|c||c|c|}
\hline
Feature for evaluation                 & CSIQ          & KADID \\ \hline
Language feature + Image feature & 0.941          & 0.939 \\ \hline
SLIQUE (only image feature)       & \textbf{0.966} & \textbf{0.957} \\ \hline
\end{tabular}

\label{features_syn}
\end{table}

\begin{table}[h]
\centering
\caption{Exploring more features used in the evaluation of authentic distortion datasets.}
\fontsize{9}{11}\selectfont
\begin{tabular}{|c||c|c|}
\hline
Feature for evaluation                 & KonIQ          & CLIVE \\ \hline
Language feature + Image feature & 0.899          & 0.861 \\ \hline
SLIQUE (only image feature)       & \textbf{0.916} & \textbf{0.897} \\ \hline
\end{tabular}

\label{features_ugc}
\end{table}

\section{Concluding Remarks}
We reasoned that the semantic contents, the distortions characteristics, and the appearance properties all affect the perceived image quality. Recognizing one of the most challenging problems in Image Quality Assessment (IQA) is that these intertwining factors confound a solution, we have developed a new blind image quality assessment (BIQA) model that exploits vision-language supervised modeling and visual self-supervised learning to acquire all these quality relevant representation features together. We have shown that our Self-supervised and Vision-Language supervised Image QUality Evaluator (SLIQUE), the first BIQA model that jointly models image semantic contents, distortions and appearances together and trained based on the newly constructed large Text
Annotated Distortion, Appearance and Content (TADAC)
image database, is capable of achieving superior performances to state of the art BIQA models, demonstrating the soundness of SLIQUE's design principles and the effectiveness of its implementation.

\bibliographystyle{IEEEtran}

\bibliography{SLIQUE}  

\begin{thebibliography}{10}
\providecommand{\url}[1]{#1}
\csname url@samestyle\endcsname
\providecommand{\newblock}{\relax}
\providecommand{\bibinfo}[2]{#2}
\providecommand{\BIBentrySTDinterwordspacing}{\spaceskip=0pt\relax}
\providecommand{\BIBentryALTinterwordstretchfactor}{4}
\providecommand{\BIBentryALTinterwordspacing}{\spaceskip=\fontdimen2\font plus
\BIBentryALTinterwordstretchfactor\fontdimen3\font minus
  \fontdimen4\font\relax}
\providecommand{\BIBforeignlanguage}[2]{{%
\expandafter\ifx\csname l@#1\endcsname\relax
\typeout{** WARNING: IEEEtran.bst: No hyphenation pattern has been}%
\typeout{** loaded for the language `#1'. Using the pattern for}%
\typeout{** the default language instead.}%
\else
\language=\csname l@#1\endcsname
\fi
#2}}
\providecommand{\BIBdecl}{\relax}
\BIBdecl

\bibitem{chen2020simple}
T.~Chen, S.~Kornblith, M.~Norouzi, and G.~Hinton, ``A simple framework for
  contrastive learning of visual representations,'' in \emph{International
  Conference on Machine Learning}.\hskip 1em plus 0.5em minus 0.4em\relax PMLR,
  2020, pp. 1597--1607.

\bibitem{he2020momentum}
K.~He, H.~Fan, Y.~Wu, S.~Xie, and R.~Girshick, ``Momentum contrast for
  unsupervised visual representation learning,'' in \emph{Proceedings of the
  IEEE/CVF Conference on Computer Vision and Pattern Recognition}, 2020, pp.
  9729--9738.

\bibitem{ijcai2023p188}
Y.~Yuan, X.~Fu, Y.~Yu, and X.~Li, ``Densedino: Boosting dense self-supervised
  learning with token-based point-level consistency,'' in \emph{Proceedings of
  the Thirty-Second International Joint Conference on Artificial Intelligence,
  {IJCAI-23}}, 2023, pp. 1695--1703.

\bibitem{synmos}
S.~Athar, Z.~Wang, and Z.~Wang, ``Deep neural networks for blind image quality
  assessment: addressing the data challenge,'' \emph{arXiv preprint
  arXiv:2109.12161}, 2021.

\bibitem{su2020blindly}
S.~Su, Q.~Yan, Y.~Zhu, C.~Zhang, X.~Ge, J.~Sun, and Y.~Zhang, ``Blindly assess
  image quality in the wild guided by a self-adaptive hyper network,'' in
  \emph{Proceedings of the IEEE/CVF Conference on Computer Vision and Pattern
  Recognition}, 2020, pp. 3667--3676.

\bibitem{topiq}
C.~Chen, J.~Mo, J.~Hou, H.~Wu, L.~Liao, W.~Sun, Q.~Yan, and W.~Lin, ``Topiq: A
  top-down approach from semantics to distortions for image quality
  assessment,'' \emph{IEEE Transactions on Image Processing}, 2024.

\bibitem{mittal2012making}
A.~Mittal, R.~Soundararajan, and A.~C. Bovik, ``Making a “completely blind”
  image quality analyzer,'' \emph{IEEE Signal Processing Letters}, vol.~20,
  no.~3, pp. 209--212, 2012.

\bibitem{ying2020patches}
Z.~Ying, H.~Niu, P.~Gupta, D.~Mahajan, D.~Ghadiyaram, and A.~Bovik, ``From
  patches to pictures (paq-2-piq): Mapping the perceptual space of picture
  quality,'' in \emph{Proceedings of the IEEE/CVF Conference on Computer Vision
  and Pattern Recognition}, 2020, pp. 3575--3585.

\bibitem{ke2021musiq}
J.~Ke, Q.~Wang, Y.~Wang, P.~Milanfar, and F.~Yang, ``Musiq: Multi-scale image
  quality transformer,'' in \emph{Proceedings of the IEEE/CVF International
  Conference on Computer Vision}, 2021, pp. 5148--5157.

\bibitem{bosse2017deep}
S.~Bosse, D.~Maniry, K.-R. M{\"u}ller, T.~Wiegand, and W.~Samek, ``Deep neural
  networks for no-reference and full-reference image quality assessment,''
  \emph{IEEE Transactions on Image Processing}, vol.~27, no.~1, pp. 206--219,
  2017.

\bibitem{zhu2020metaiqa}
H.~Zhu, L.~Li, J.~Wu, W.~Dong, and G.~Shi, ``Metaiqa: Deep meta-learning for
  no-reference image quality assessment,'' in \emph{Proceedings of the IEEE/CVF
  Conference on Computer Vision and Pattern Recognition}, 2020, pp.
  14\,143--14\,152.

\bibitem{zhang2018blind}
W.~Zhang, K.~Ma, J.~Yan, D.~Deng, and Z.~Wang, ``Blind image quality assessment
  using a deep bilinear convolutional neural network,'' \emph{IEEE Transactions
  on Circuits and Systems for Video Technology}, vol.~30, no.~1, pp. 36--47,
  2018.

\bibitem{sun2023blind}
W.~Sun, X.~Min, D.~Tu, S.~Ma, and G.~Zhai, ``Blind quality assessment for
  in-the-wild images via hierarchical feature fusion and iterative mixed
  database training,'' \emph{IEEE Journal of Selected Topics in Signal
  Processing}, 2023.

\bibitem{wang2023blind}
H.~Wang, G.~Wang, W.~Xia, Z.~Yang, H.~Yu, L.~Fang, and Y.~Zhang, ``Blind image
  quality assessment via deep response feature decomposition and aggregation,''
  \emph{IEEE Journal of Selected Topics in Signal Processing}, 2023.

\bibitem{zhang2023blind}
W.~Zhang, G.~Zhai, Y.~Wei, X.~Yang, and K.~Ma, ``Blind image quality assessment
  via vision-language correspondence: A multitask learning perspective,'' in
  \emph{Proceedings of the IEEE/CVF Conference on Computer Vision and Pattern
  Recognition}, 2023, pp. 14\,071--14\,081.

\bibitem{radford2019language}
A.~Radford, J.~Wu, R.~Child, D.~Luan, D.~Amodei, I.~Sutskever \emph{et~al.},
  ``Language models are unsupervised multitask learners,'' \emph{OpenAI blog},
  vol.~1, no.~8, p.~9, 2019.

\bibitem{srinath2024learning}
S.~Srinath, S.~Mitra, S.~Rao, and R.~Soundararajan, ``Learning generalizable
  perceptual representations for data-efficient no-reference image quality
  assessment,'' in \emph{Proceedings of the IEEE/CVF Winter Conference on
  Applications of Computer Vision}, 2024, pp. 22--31.

\bibitem{gabbay2021image}
A.~Gabbay, N.~Cohen, and Y.~Hoshen, ``An image is worth more than a thousand
  words: Towards disentanglement in the wild,'' \emph{Advances in Neural
  Information Processing Systems}, vol.~34, pp. 9216--9228, 2021.

\bibitem{xu2022predict}
Z.~Xu, T.~Lin, H.~Tang, F.~Li, D.~He, N.~Sebe, R.~Timofte, L.~Van~Gool, and
  E.~Ding, ``Predict, prevent, and evaluate: Disentangled text-driven image
  manipulation empowered by pre-trained vision-language model,'' in
  \emph{Proceedings of the IEEE/CVF Conference on Computer Vision and Pattern
  Recognition}, 2022, pp. 18\,229--18\,238.

\bibitem{hessel2021clipscore}
J.~Hessel, A.~Holtzman, M.~Forbes, R.~L. Bras, and Y.~Choi, ``Clipscore: A
  reference-free evaluation metric for image captioning,'' \emph{arXiv preprint
  arXiv:2104.08718}, 2021.

\bibitem{zhong2022regionclip}
Y.~Zhong, J.~Yang, P.~Zhang, C.~Li, N.~Codella, L.~H. Li, L.~Zhou, X.~Dai,
  L.~Yuan, Y.~Li \emph{et~al.}, ``Regionclip: Region-based language-image
  pretraining,'' in \emph{Proceedings of the IEEE/CVF Conference on Computer
  Vision and Pattern Recognition}, 2022, pp. 16\,793--16\,803.

\bibitem{shi2022proposalclip}
H.~Shi, M.~Hayat, Y.~Wu, and J.~Cai, ``Proposalclip: Unsupervised open-category
  object proposal generation via exploiting clip cues,'' in \emph{Proceedings
  of the IEEE/CVF Conference on Computer Vision and Pattern Recognition}, 2022,
  pp. 9611--9620.

\bibitem{rao2022denseclip}
Y.~Rao, W.~Zhao, G.~Chen, Y.~Tang, Z.~Zhu, G.~Huang, J.~Zhou, and J.~Lu,
  ``Denseclip: Language-guided dense prediction with context-aware prompting,''
  in \emph{Proceedings of the IEEE/CVF Conference on Computer Vision and
  Pattern Recognition}, 2022, pp. 18\,082--18\,091.

\bibitem{zhou2022extract}
C.~Zhou, C.~C. Loy, and B.~Dai, ``Extract free dense labels from clip,'' in
  \emph{European Conference on Computer Vision}.\hskip 1em plus 0.5em minus
  0.4em\relax Springer, 2022, pp. 696--712.

\bibitem{yang2022vision}
J.~Yang, J.~Duan, S.~Tran, Y.~Xu, S.~Chanda, L.~Chen, B.~Zeng, T.~Chilimbi, and
  J.~Huang, ``Vision-language pre-training with triple contrastive learning,''
  in \emph{Proceedings of the IEEE/CVF Conference on Computer Vision and
  Pattern Recognition}, 2022, pp. 15\,671--15\,680.

\bibitem{uniclip}
J.~Lee, J.~Kim, H.~Shon, B.~Kim, S.~H. Kim, H.~Lee, and J.~Kim, ``Uniclip:
  Unified framework for contrastive language-image pre-training,''
  \emph{Advances in Neural Information Processing Systems}, vol.~35, pp.
  1008--1019, 2022.

\bibitem{wang2023exploring}
J.~Wang, K.~C. Chan, and C.~C. Loy, ``Exploring clip for assessing the look and
  feel of images,'' in \emph{Proceedings of the AAAI Conference on Artificial
  Intelligence}, vol.~37, no.~2, 2023, pp. 2555--2563.

\bibitem{radford2021learning}
A.~Radford, J.~W. Kim, C.~Hallacy, A.~Ramesh, G.~Goh, S.~Agarwal, G.~Sastry,
  A.~Askell, P.~Mishkin, J.~Clark \emph{et~al.}, ``Learning transferable visual
  models from natural language supervision,'' in \emph{International Conference
  on Machine Learning}.\hskip 1em plus 0.5em minus 0.4em\relax PMLR, 2021, pp.
  8748--8763.

\bibitem{He_2022_CVPR}
K.~He, X.~Chen, S.~Xie, Y.~Li, P.~Doll\'ar, and R.~Girshick, ``Masked
  autoencoders are scalable vision learners,'' in \emph{Proceedings of the
  IEEE/CVF Conference on Computer Vision and Pattern Recognition}, 2022.

\bibitem{10196496}
Z.~Zhou, F.~Zhou, and G.~Qiu, ``Blind image quality assessment based on
  separate representations and adaptive interaction of content and
  distortion,'' \emph{IEEE Transactions on Circuits and Systems for Video
  Technology}, vol. Early Access, pp. 1--14, 2023.

\bibitem{madhusudana2022image}
P.~C. Madhusudana, N.~Birkbeck, Y.~Wang, B.~Adsumilli, and A.~C. Bovik, ``Image
  quality assessment using contrastive learning,'' \emph{IEEE Transactions on
  Image Processing}, vol.~31, pp. 4149--4161, 2022.

\bibitem{saha2023re}
A.~Saha, S.~Mishra, and A.~C. Bovik, ``Re-iqa: Unsupervised learning for image
  quality assessment in the wild,'' in \emph{Proceedings of the IEEE/CVF
  Conference on Computer Vision and Pattern Recognition}, 2023, pp. 5846--5855.

\bibitem{deng2009imagenet}
J.~Deng, W.~Dong, R.~Socher, L.-J. Li, K.~Li, and L.~Fei-Fei, ``Imagenet: A
  large-scale hierarchical image database,'' in \emph{Proceedings of the
  IEEE/CVF Conference on Computer Vision and Pattern Recognition}, 2009, pp.
  248--255.

\bibitem{chen2022spiq}
P.~Chen, L.~Li, Q.~Wu, and J.~Wu, ``Spiq: A self-supervised pre-trained model
  for image quality assessment,'' \emph{IEEE Signal Processing Letters},
  vol.~29, pp. 513--517, 2022.

\bibitem{Zhao_2023_CVPR}
K.~Zhao, K.~Yuan, M.~Sun, M.~Li, and X.~Wen, ``Quality-aware pre-trained models
  for blind image quality assessment,'' in \emph{Proceedings of the IEEE/CVF
  Conference on Computer Vision and Pattern Recognition}, June 2023, pp.
  22\,302--22\,313.

\bibitem{oord2018representation}
A.~v.~d. Oord, Y.~Li, and O.~Vinyals, ``Representation learning with
  contrastive predictive coding,'' \emph{arXiv preprint arXiv:1807.03748},
  2018.

\bibitem{9171494}
F.~Zhou, R.~Yao, G.~Liao, B.~Liu, and G.~Qiu, ``Visual saliency via embedding
  hierarchical knowledge in a deep neural network,'' \emph{IEEE Transactions on
  Image Processing}, vol.~29, pp. 8490--8505, 2020.

\bibitem{mu2022slip}
N.~Mu, A.~Kirillov, D.~Wagner, and S.~Xie, ``Slip: Self-supervision meets
  language-image pre-training,'' in \emph{European Conference on Computer
  Vision}.\hskip 1em plus 0.5em minus 0.4em\relax Springer, 2022, pp. 529--544.

\bibitem{deepfl-iqa}
H.~Lin, V.~Hosu, and D.~Saupe, ``Deepfl-iqa: Weak supervision for deep iqa
  feature learning,'' \emph{arXiv preprint arXiv:2001.08113}, 2020.

\bibitem{murray2012ava}
N.~Murray, L.~Marchesotti, and F.~Perronnin, ``Ava: A large-scale database for
  aesthetic visual analysis,'' in \emph{Proceedings of the IEEE/CVF Conference
  on Computer Vision and Pattern Recognition}.\hskip 1em plus 0.5em minus
  0.4em\relax IEEE, 2012, pp. 2408--2415.

\bibitem{lin2014microsoft}
T.-Y. Lin, M.~Maire, S.~Belongie, J.~Hays, P.~Perona, D.~Ramanan,
  P.~Doll{\'a}r, and C.~L. Zitnick, ``Microsoft coco: Common objects in
  context,'' in \emph{European Conference on Computer Vision}.\hskip 1em plus
  0.5em minus 0.4em\relax Springer, 2014, pp. 740--755.

\bibitem{everingham2010pascal}
M.~Everingham, L.~Van~Gool, C.~K. Williams, J.~Winn, and A.~Zisserman, ``The
  pascal visual object classes (voc) challenge,'' \emph{International Journal
  of Computer Vision}, vol.~88, pp. 303--338, 2010.

\bibitem{zhou2017places}
B.~Zhou, A.~Lapedriza, A.~Khosla, A.~Oliva, and A.~Torralba, ``Places: A 10
  million image database for scene recognition,'' \emph{IEEE Transactions on
  Pattern Analysis and Machine Itelligence}, vol.~40, no.~6, pp. 1452--1464,
  2017.

\bibitem{mavridaki2014no}
E.~Mavridaki and V.~Mezaris, ``No-reference blur assessment in natural images
  using fourier transform and spatial pyramids,'' in \emph{2014 IEEE
  International Conference on Image Processing (ICIP)}.\hskip 1em plus 0.5em
  minus 0.4em\relax IEEE, 2014, pp. 566--570.

\bibitem{mokady2021clipcap}
R.~Mokady, A.~Hertz, and A.~H. Bermano, ``Clipcap: Clip prefix for image
  captioning,'' \emph{arXiv preprint arXiv:2111.09734}, 2021.

\bibitem{6202326}
P.~V. Vu and D.~M. Chandler, ``A fast wavelet-based algorithm for global and
  local image sharpness estimation,'' \emph{IEEE Signal Processing Letters},
  vol.~19, no.~7, pp. 423--426, 2012.

\bibitem{10.1117/12.477378}
\BIBentryALTinterwordspacing
D.~Hasler and S.~E. Suesstrunk, ``{Measuring colorfulness in natural images},''
  in \emph{Human Vision and Electronic Imaging VIII}, B.~E. Rogowitz and T.~N.
  Pappas, Eds., vol. 5007, International Society for Optics and
  Photonics.\hskip 1em plus 0.5em minus 0.4em\relax SPIE, 2003, pp. 87 -- 95.
  [Online]. Available: \url{https://doi.org/10.1117/12.477378}
\BIBentrySTDinterwordspacing

\bibitem{gpt4}
J.~Achiam, S.~Adler, S.~Agarwal, L.~Ahmad, I.~Akkaya, F.~L. Aleman, D.~Almeida,
  J.~Altenschmidt, S.~Altman, S.~Anadkat \emph{et~al.}, ``Gpt-4 technical
  report,'' \emph{arXiv preprint arXiv:2303.08774}, 2023.

\bibitem{gptjudging}
L.~Zheng, W.-L. Chiang, Y.~Sheng, S.~Zhuang, Z.~Wu, Y.~Zhuang, Z.~Lin, Z.~Li,
  D.~Li, E.~Xing \emph{et~al.}, ``Judging llm-as-a-judge with mt-bench and
  chatbot arena,'' \emph{Advances in Neural Information Processing Systems},
  vol.~36, pp. 46\,595--46\,623, 2023.

\bibitem{sheikh2006statistical}
H.~R. Sheikh, M.~F. Sabir, and A.~C. Bovik, ``A statistical evaluation of
  recent full reference image quality assessment algorithms,'' \emph{IEEE
  Transactions on Image Processing}, vol.~15, no.~11, pp. 3440--3451, 2006.

\bibitem{larson2010most}
E.~C. Larson and D.~M. Chandler, ``Most apparent distortion: full-reference
  image quality assessment and the role of strategy,'' \emph{Journal of
  Electronic Imaging}, vol.~19, no.~1, pp. 011\,006--011\,006, 2010.

\bibitem{ponomarenko2015image}
N.~Ponomarenko, L.~Jin, O.~Ieremeiev, V.~Lukin, K.~Egiazarian, J.~Astola,
  B.~Vozel, K.~Chehdi, M.~Carli, F.~Battisti \emph{et~al.}, ``Image database
  tid2013: Peculiarities, results and perspectives,'' \emph{Signal Processing:
  Image Communication}, vol.~30, pp. 57--77, 2015.

\bibitem{lin2019kadid}
H.~Lin, V.~Hosu, and D.~Saupe, ``Kadid-10k: A large-scale artificially
  distorted iqa database,'' in \emph{2019 Eleventh International Conference on
  Quality of Multimedia Experience (QoMEX)}.\hskip 1em plus 0.5em minus
  0.4em\relax IEEE, 2019, pp. 1--3.

\bibitem{hosu2020koniq}
V.~Hosu, H.~Lin, T.~Sziranyi, and D.~Saupe, ``Koniq-10k: An ecologically valid
  database for deep learning of blind image quality assessment,'' \emph{IEEE
  Transactions on Image Processing}, vol.~29, pp. 4041--4056, 2020.

\bibitem{ghadiyaram2015massive}
D.~Ghadiyaram and A.~C. Bovik, ``Massive online crowdsourced study of
  subjective and objective picture quality,'' \emph{IEEE Transactions on Image
  Processing}, vol.~25, no.~1, pp. 372--387, 2015.

\bibitem{fang2020perceptual}
Y.~Fang, H.~Zhu, Y.~Zeng, K.~Ma, and Z.~Wang, ``Perceptual quality assessment
  of smartphone photography,'' in \emph{Proceedings of the IEEE/CVF Conference
  on Computer Vision and Pattern Recognition}, 2020, pp. 3677--3686.

\bibitem{mittal2012no}
A.~Mittal, A.~K. Moorthy, and A.~C. Bovik, ``No-reference image quality
  assessment in the spatial domain,'' \emph{IEEE Transactions on Image
  Processing}, vol.~21, no.~12, pp. 4695--4708, 2012.

\bibitem{ye2012unsupervised}
P.~Ye, J.~Kumar, L.~Kang, and D.~Doermann, ``Unsupervised feature learning
  framework for no-reference image quality assessment,'' in \emph{Proceedings
  of the IEEE/CVF Conference on Computer Vision and Pattern Recognition}.\hskip
  1em plus 0.5em minus 0.4em\relax IEEE, 2012, pp. 1098--1105.

\bibitem{zeng2017probabilistic}
H.~Zeng, L.~Zhang, and A.~C. Bovik, ``A probabilistic quality representation
  approach to deep blind image quality prediction,'' \emph{arXiv preprint
  arXiv:1708.08190}, 2017.

\bibitem{zhang2021uncertainty}
W.~Zhang, K.~Ma, G.~Zhai, and X.~Yang, ``Uncertainty-aware blind image quality
  assessment in the laboratory and wild,'' \emph{IEEE Transactions on Image
  Processing}, vol.~30, pp. 3474--3486, 2021.

\bibitem{uif}
Y.~Zheng, W.~Chen, R.~Lin, T.~Zhao, and P.~Le~Callet, ``Uif: An objective
  quality assessment for underwater image enhancement,'' \emph{IEEE
  Transactions on Image Processing}, vol.~31, pp. 5456--5468, 2022.

\bibitem{uid2021}
G.~Hou, Y.~Li, H.~Yang, K.~Li, and Z.~Pan, ``Uid2021: An underwater image
  dataset for evaluation of no-reference quality assessment metrics,''
  \emph{ACM Transactions on Multimedia Computing, Communications and
  Applications}, vol.~19, no.~4, pp. 1--24, 2023.

\bibitem{QADS}
F.~Zhou, R.~Yao, B.~Liu, and G.~Qiu, ``Visual quality assessment for
  super-resolved images: Database and method,'' \emph{IEEE Transactions on
  Image Processing}, vol.~28, no.~7, pp. 3528--3541, 2019.

\bibitem{faceiqa}
S.~Su, H.~Lin, V.~Hosu, O.~Wiedemann, J.~Sun, Y.~Zhu, H.~Liu, Y.~Zhang, and
  D.~Saupe, ``Going the extra mile in face image quality assessment: A novel
  database and model,'' \emph{IEEE Transactions on Multimedia}, 2023.

\bibitem{qins}
H.~Wu, Z.~Zhang, E.~Zhang, C.~Chen, L.~Liao, A.~Wang, K.~Xu, C.~Li, J.~Hou,
  G.~Zhai \emph{et~al.}, ``Q-instruct: Improving low-level visual abilities for
  multi-modality foundation models,'' in \emph{Proceedings of the IEEE/CVF
  Conference on Computer Vision and Pattern Recognition}, 2024, pp.
  25\,490--25\,500.

\end{thebibliography}

\vspace{11pt}
\vspace{-33pt}
\begin{IEEEbiography}[{\includegraphics[width=1in,height=1.25in,clip,keepaspectratio]{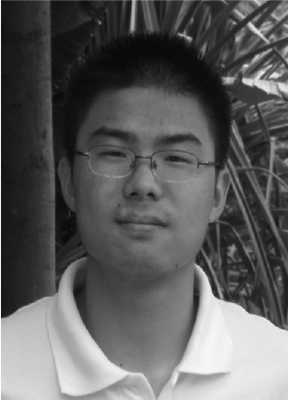}}]{Fei Zhou}
received the B.Eng. degree in electronics and information engineering from the Huazhong University of Science and Technology, in 2007, and the Ph.D. degree in electronic engineering, Tsinghua University, in 2013. From 2013 to 2016, he was a post-doctoral fellow with the Graduate School at Shenzhen, Tsinghua University. From 2017 to 2018, he was a Visiting Scholar with the Department of Statistical Science, University College London. He is currently an Associate Professor with the College of Electronic and Information Engineering, Shenzhen University. He has authored over 60 papers internationally. His research interests include image super-resolution, image decomposition, image quality assessment and inverse tone mapping. He is a reviewer of many well-known journals, including IEEE TIP, IEEE TCSVT, Information Sciences, etc.
\end{IEEEbiography}

\vspace{11pt}
\vspace{-33pt}
\begin{IEEEbiography}[{\includegraphics[width=1in,clip,keepaspectratio]{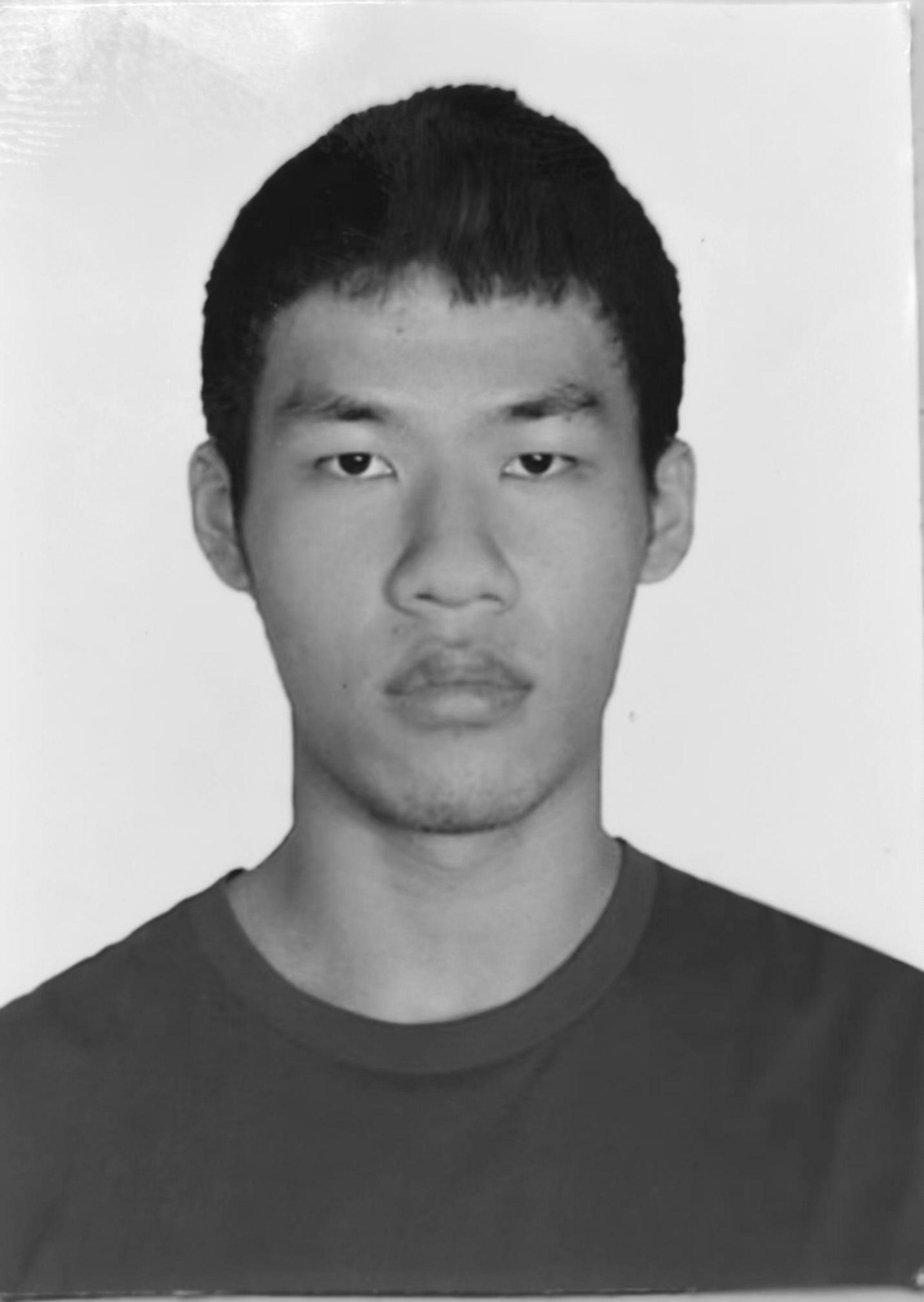}}]{Tianhao Gu}
received the B.Eng. degree from Shenzhen University, Shenzhen, China, in 2023, where he is currently pursuing the master's degree with the College of Electronic and Information Engineering. His research focus is image quality assessment.
\end{IEEEbiography}

\vspace{11pt}
\vspace{-33pt}
\begin{IEEEbiography}[{\includegraphics[width=1in,clip,keepaspectratio]{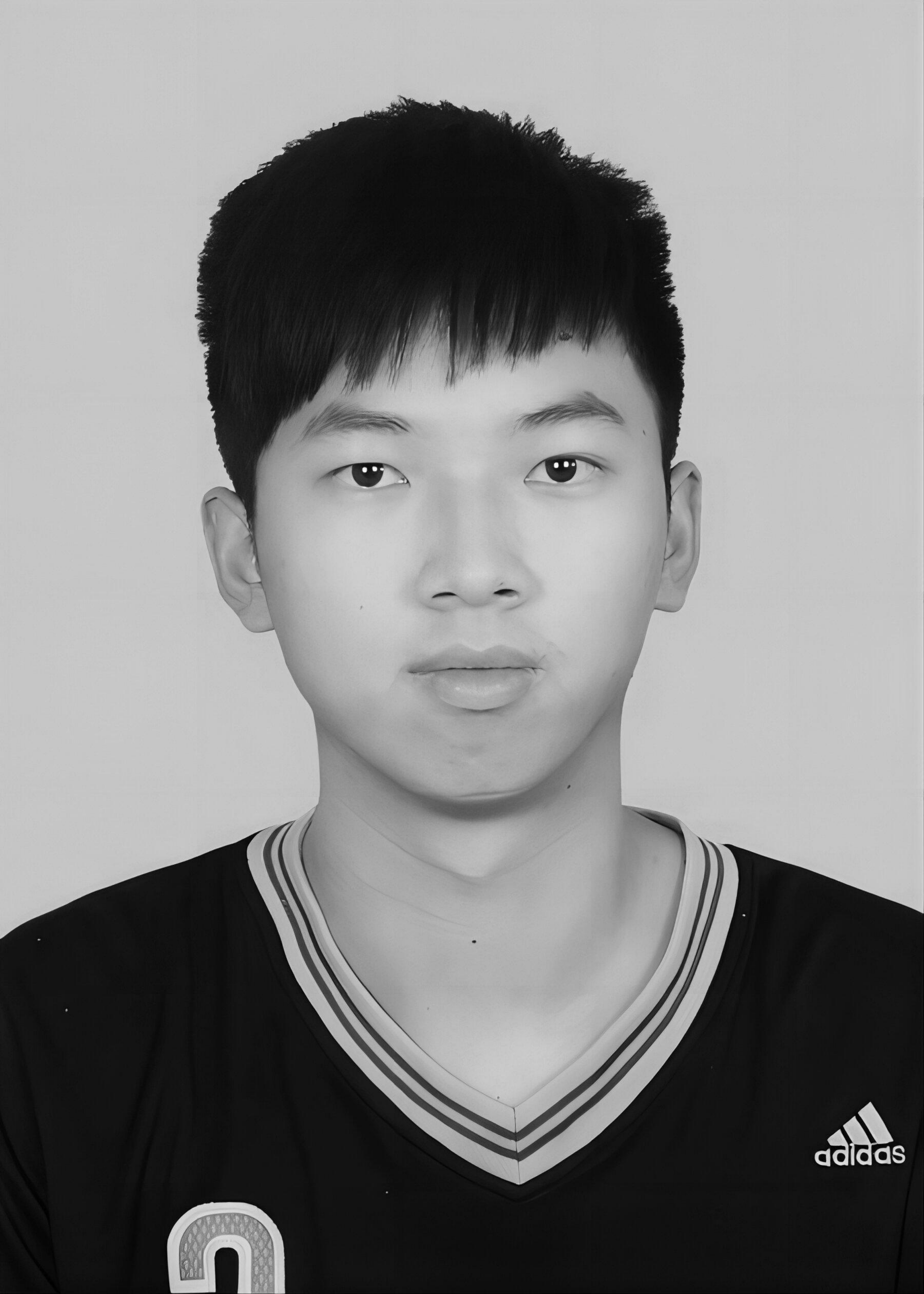}}]{Zhicong Huang}
received the B.Eng. degree from Tianjin Polytechnic University, Tianjin, China, in 2020, and is currently studying for a master's degree in the College of Electronic and Information Engineering of Shenzhen University, Shenzhen, China. His research interests include image quality assessment.
\end{IEEEbiography}

\vspace{11pt}
\vspace{-33pt}
\begin{IEEEbiography}[{\includegraphics[width=1in,height=1.25in,clip,keepaspectratio]{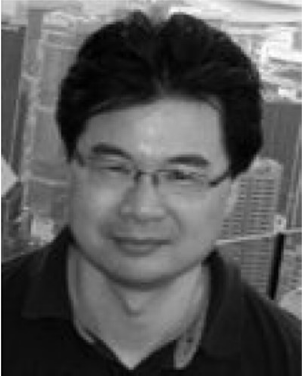}}]{Guoping Qiu}
is a Distinguished Professor of Information Engineering, Director of Shenzhen University Intelligent Robotics Centre at Shenzhen University, China, and a Chair Professor of Visual Information Processing at the University of Nottingham, Nottingham, UK. He has taught in universities in the UK and Hong Kong and also consulted for multinational companies in Europe, Hong Kong and China. His research interests include image processing, pattern recognition, and machine learning. He is particularly known for his pioneering research in high dynamic range imaging and machine learning based image processing technologies. He has published widely and holds several European and US patents. Technologies developed in his lab have laid the cornerstone for successful spinout companies who are developing advanced digital photography software enjoyed by tens of millions of global users.
\end{IEEEbiography}

\end{document}